# Optical Script Identification for multi-lingual Indic-script

Sidhantha Poddar
Vellore Instiiute of Technology
Vellore India
https://orcid.org/0000-0002-4936-7060

Rohan Gupta
Vellore Instiiute of Technology
Vellore India
rohangpt583@gmail.com

***Abstract*—Script identification and text recognition are some of the major domains in the application of Artificial Intelligence. In this era of digitalization, the use of digital note-taking has become a common practice. Still, conventional methods of using pen and paper is a prominent way of writing. This leads to the classification of scripts based on the method they are obtained. A survey on the current methodologies and state-of-art methods used for processing and identification would prove beneficial for researchers. The aim of this article is to discuss the advancement in the techniques for script pre-processing and text recognition. In India there are twelve prominent Indic scripts, unlike the English language, these scripts have layers of characteristics. Complex characteristics such as similarity in text shape make them difficult to recognize and analyze, thus this requires advance preprocessing methods for their accurate recognition. A sincere attempt is made in this survey to provide a comparison between all algorithms. We hope that this survey would provide insight to a researcher working not only on Indic scripts but also other languages.**

## I. INTRODUCTION (*HEADING 1*)

Over the past few decades, the world has seen a great advancement in technology. Technology has developed in all possible sectors be it processing power or advance optimized computational algorithm. In the previous few years, there has been massive development in fields like Machine Learning, Artificial Intelligence, Robotics, Internet of Things, Cloud Computing. These new areas of, human to machine and machine to world interaction have opened numerous applications for the use of incredible processing power and the massive resources.

The field of artificial intelligence has a major impact on the way of living a normal person leads. Now AI – Artificial Intelligence can be categorized into many parts viz. Natural Language Processing, Deep Learning, Machine Learning, Expert Systems, Image Processing being the most prominent of these. Now all these fields are inter-dependent and used in parallel for different phases.

Natural Language processing is the part of AI which allows the machine to interpret human languages, be it spoken or written. Human speaks, write even think in different languages, our brain functions in a very complex manner and can produce infinite possible ways to express a thought. We have different personalities as well as different ways of expressing a language. Even the same written text can have multiple forms. These infinite variations cannot be ruled or cased for all persons. Thus, we need a solution so that machines can interact efficiently with us. Here advance algorithms of AI and NLP come to the picture. Instead of accounting for the rules provided by the System as default, these algorithms search characteristics in the data, check for similarities between data and form its own set of rules to use them in real-time.

One of the applications of NLP is Character Recognition. Humans tend to understand scripts written in different handwriting and in different languages (provided that they know the language). But when it comes to computers, optical recognition is still under development phase and is not able to recognize human written scripts very accurately and efficiently. The complex nature of curves and variations in handwriting are not accounted for properly for commercial systems. This is a very crucial application in AI and computer vision. Text recognition and computer vision can be used in various sectors of work like preserving old documents, finding and searching in large sets of scanned scripts, which would otherwise cost a lot of resources and require exponentially greater time.

In India, there are 72 recognized spoken languages out of which 12 are prominently written scripts. Unlike English or other languages, they aren’t composed of simple strokes or simple curves but are set of complex symbols that have been placed in combination with one another. Example in Devanagari scripts we have base letter combined with vowels which act as a modifier; these set of combinations can produce an infinite set of the word thus making it hard to recognize; one possible solution is to segment the words and into parts and recognize them separately. Indian cultures are among the oldest culture and Indic scripts being derived from Sanskrit can be analyzed; their recognition can lead to the new discovery of methods of recognition of older un-recognized/documented scripts. Those scripts could give an insight into human evolution.

This paper provides an insight into the various method for preprocessing Indic scripts and methods that could be used to recognize characters of the script. It gives a comparative analysis of datasets along with the training model of the recognition system. This survey could be used by researchers to get a head start for their work on recognition of Indic scripts

Handwritten Character recognition has been proven useful in Its application since the amount of manual labor it takes to read and understand each document is staggering. With Handwritten character recognition, the documents can be easily be converted to digital form making the overall system more efficient. But with traditional machine learning algorithms, the amount of resources and time it takes to train and build a model is a lot. Also, the performance of the model reaches a constant value after a certain amount of data [7].

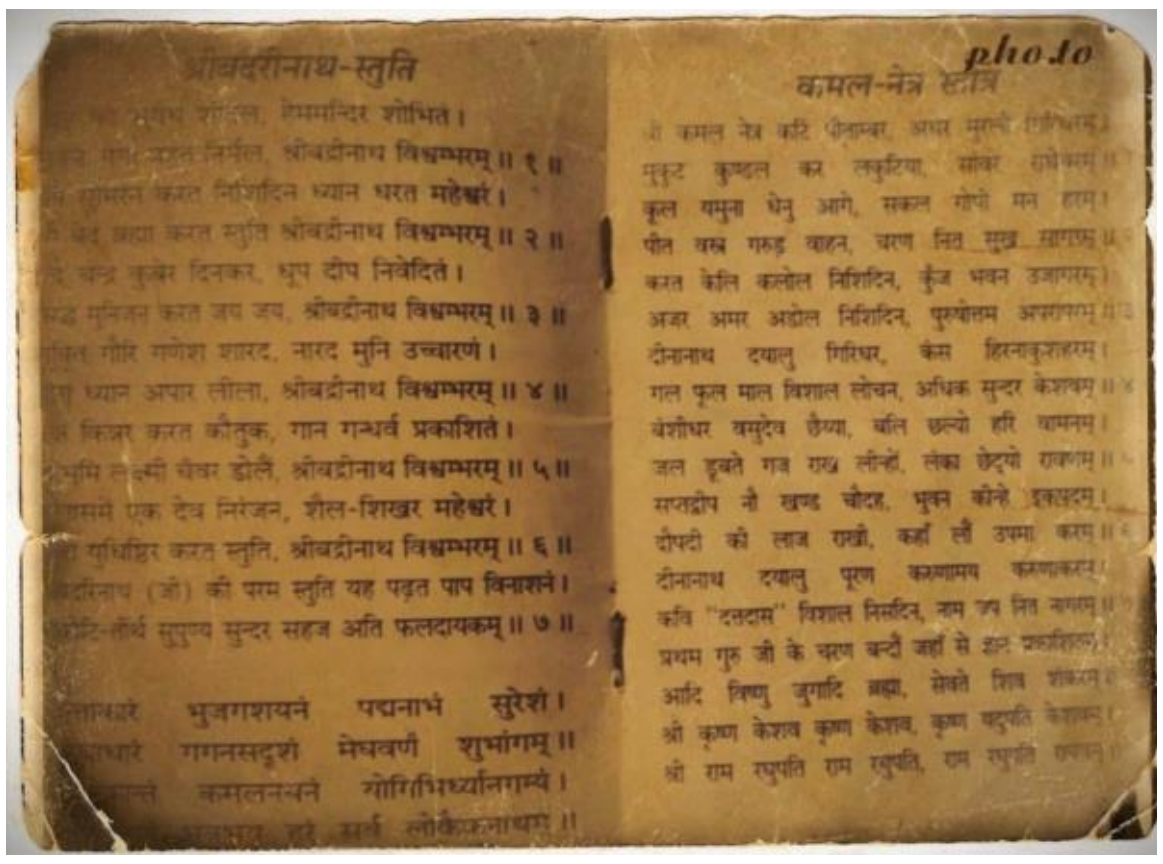

*Figure 2.1 - Sample preserved religious Indic script*

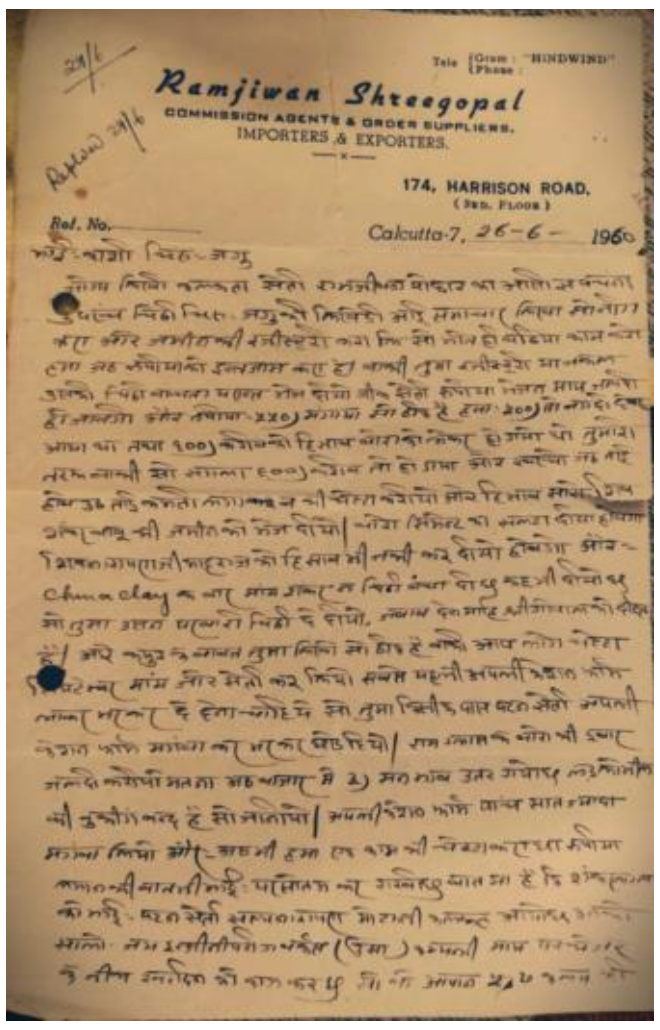

*Figure 1.2 – Preserved Official Letter form 1960*

Even though today's generation is shifting towards a paperless world, there are still several scenarios in which handwritten communications is preferred. For example, conventional pen and paper or modern-day Apple pencil and iPad. The handwritten text offers more convenience, resourcefulness, and efficiency. Numerous documents such as paying invoices and question papers often consist of blank fields that need to be converted to digital form [1].

Various studies have been conducted in the past to study the techniques used for recognizing visually similar characters. The existing methods can be classified into two categories [.2]:

i. Classifier Centric- This approach gives more emphasis on developing a complicated and highly distinguishing classifier in order to achieve better discriminating similar characters. Although classifiers proposed in various studies demonstrated good performance, they usually required large samples for training purposes.
ii. Feature Centric – This approach is aimed at designing good quality features which can be used as an input for generic classifiers such as Modified Quadratic Discriminant Function

The obtainable performance of the handwriting recognition process is dependent on which feature set is selected for representing the input samples. Due to this a large variety of feature sets have been developed each with a different number of attributes. However, utilizing a large quantity of features may degrade the performance of learning algorithms, especially when irrelevant and recurring features are present. Therefore, a variety of feature selection methods are used for identifying the desired subset of features from the available samples. This process may consume a lot of time and therefore greedy search algorithms are used. [3]

India is a country is where many languages are spoken and written. A considerable amount of research has already been conducted in the domain of online handwritten character recognition for various Indian scripts such as Devanagari, Bengali, and Tamil scripts. Online recognition of handwritten Indic scripts is more challenging in comparison to the Latin script due to the presence of many symbols. [4]

Handwritten documents can be very difficult to work with and manual management of them consumes a lot of time and labor. In addition to this, the chances of documents being lost or damaged can incur additional losses. To overcome this problem OCR /HCR is used, which efficiently digitalizes the document. The digitalized document can then be easily stored and managed. [8]

The introduction of convolutional deep architectures has allowed improvement in the performance of the conventional feature-based approaches by great amounts. Handwritten document recognition is an extremely important task in developing a multilingual handwriting recognition system where multiple scripts might be present.

### A. Related work

Handwritten text identification using offline approach involves conventionally obtaining the input features from the image data. In the next step a classifier like Artificial Neural Network (ANN) or Gaussian Mixture Model (GMM), is used for evaluation of posterior probabilities. These calculated probabilities are supplied as an input to a Hidden Markov Model (HMM) to produce transcriptions [11]. Some of the limitations of HMMs is that they are not very efficient in simulating long term dependencies in input data stream. However, RNNs such as LSTM units can contribute towards providing a solution for these limitations. LSTMs have demonstrated extraordinary capabilities in sequence learning tasks like speech recognition, machine translation, video summarization etc. [12]

Provided a two-dimensional image, a simple method for offline identification would be to consider each column of an image as a one-dimensional vector and provide it as an input to a RNN. Wigington et al. makes use of a region proposal network to discover areas of handwritten text on a paper and then uses a line-follower algorithm to trace the hand-writing.[13]. An offline English handwriting detection approach using RNN-HMM. To achieve frame-wise labeling, they applied HMM to the training data. The frames were utilized as an input to an RNN, with matching target labels. The system was trained to get posterior probabilities that produced emission probabilities for an HMM. [14].Almazan et al. in his research work utilized word attributes such as bigrams and trigrams to feed a pyramid histogram of characters. [15]

A general word identification technique for the identification of handwritten city names in Bangla script. Gradient-based features using histogram of Oriented Gradients (HOG) feature descriptor are extracted from each of the grids. [16]A new concept introduced for image recognition tasks is Histogram Intersection. However, this method is only applicable for binary strings such as color histograms which are plotted on equally sized images. [17] Attempt to recognize/distinguish between similar-looking printed Gujarati language characters. A new mechanism called ESLPP (extended version of Supervised Locality Preserving Projection) is introduced for the same.[18]

In the field of handwriting recognition, the large variation between handwritings of different writers provide a big challenge for feature selection.[19]. The accuracy of Handwritten Chinese Character recognition was improved by using linear discriminant analysis (LDA) technique for distinguishing between similar characters. The LDA-based method is an extension of the previous compound Mahala Nobis function (CMF).[20]

The process of searching for the required feature set must traverse a search space, whose size depends on the total number of available features. This is the reason why many heuristic/greedy algorithms for finding optimal solutions have been proposed in the literature.[21]

Genetic Algorithms (GA) or Particle Swarm Optimization (PSO) have proven themselves to be efficient in searching for -optimal solutions in complex and non-linear search spaces. [22]. Pattern recognition is a vast field that has seen significant advances over the years. As the datasets under consideration grow larger and more comprehensive, using efficient techniques to process them becomes increasingly important. Research presenting a versatile technique for feature selection and extraction. [23]A feature cluster taxonomy-based technique for selecting the required feature set out of the given alternatives. [24]

A large number of studies have been using different models/algorithms – SVM[61], Use of random Forest for the OCR [62] , Used KNN (K nearest Neighbor) algorithm, which has given decent results on large datasets.[63], Used Only Convolutional Neural network for the OCR. Although it provides remarkable accuracy the cost to put it up is very high.[64]. Many authors have tried to apply different algorithms on handwritten Devanagari numeral dataset.

## B. Definition of Terms

### 1) Neural Networks

A neural network is a sequence of algorithms that aims at recognizing the underlying relationships in a given data set through a process which is similar to the way a human brain function. A neural network can adapt to changing input, so the network generates the best possible results without modifying or redefining the output criteria. The concept of a neural network can trace back its roots to artificial intelligence and is becoming widely popular.

The basic element of a neural network is the neuron. A neuron is a mathematical function that can collect and classify information according to the corresponding architecture. The network has the same characteristics as statistical methods such as curve fitting and regression analysis.

Neural networks are the most widely used in financial operations.

### 2) Long Short-Term Memory (LSTM) Networks

Long Short-Term Memory Networks (LSTM) is a special category of Recurrent Neural Networks (RNN) that can learn long term dependencies. They have shown tremendous capabilities while working on a large variety of problems and are widely popular.

Remembering the stored information for long periods is their natural behavior and not something they need to be trained for. LSTMs have a chain-like structure and instead of comprising a single neural network layer they have four and each layer interacts with another one uniquely.

Information flow in LSTMs happens with the help of a cell state. The cell state behaves like a conveyor belt with only minute interactions. It is extremely easy for information just to flow along this belt without being changed. The Cell state also comprises of gates to decide which component should pass through and which should be rejected.

### 3) Bidirectional Long Short-Term Memory Networks (BLSTMs)

Bidirectional LSTMs are an extension of the traditional LSTM model which showed increased performance for sequence classifier problems. BLSTMs involves the training of two LSTMs on the input sequence. The first LSTM is trained on the normal input sequence and the second one is trained on the reversed input sequence.

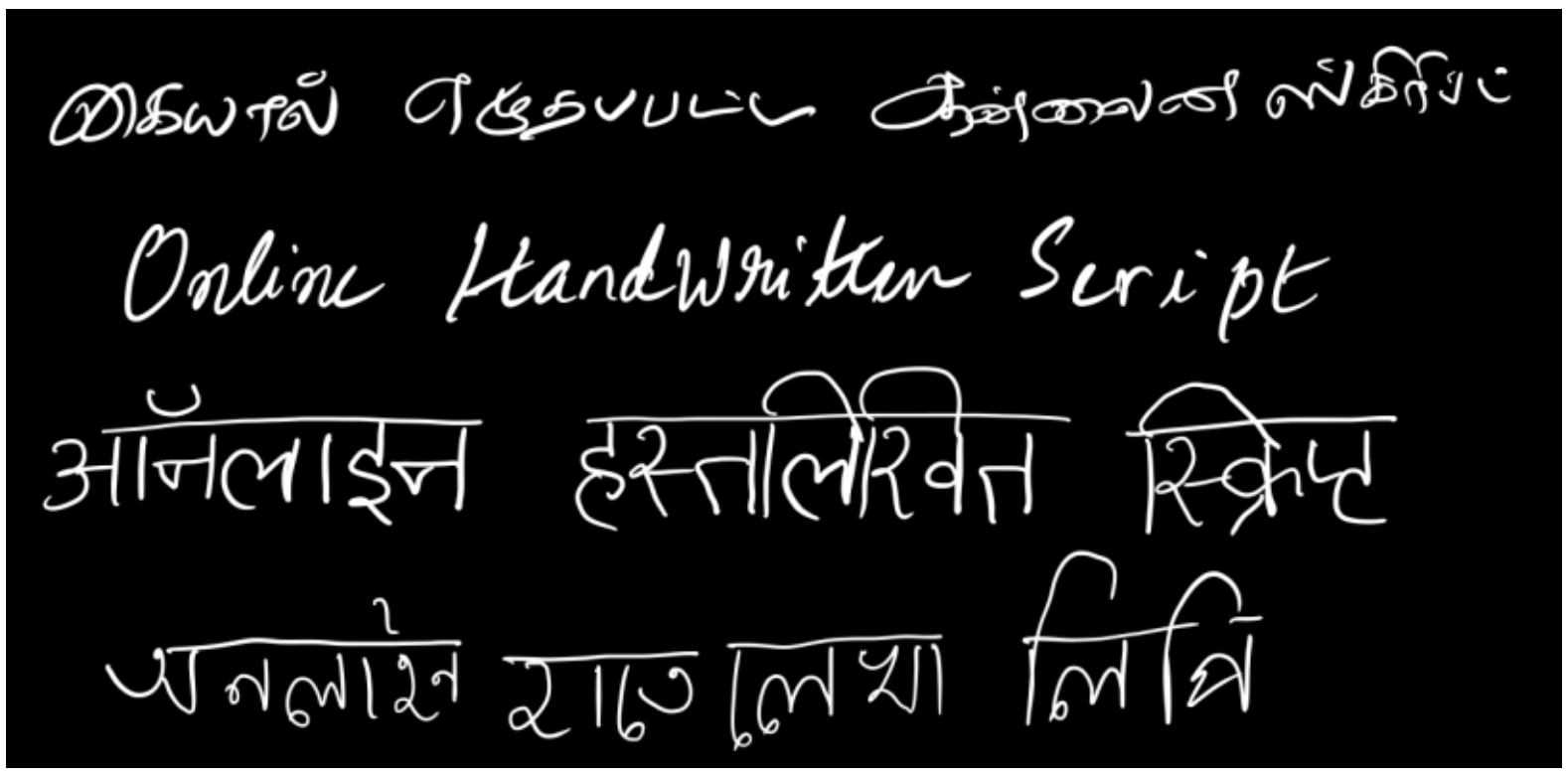

***Figure 5.1 – Handwritten online Script***

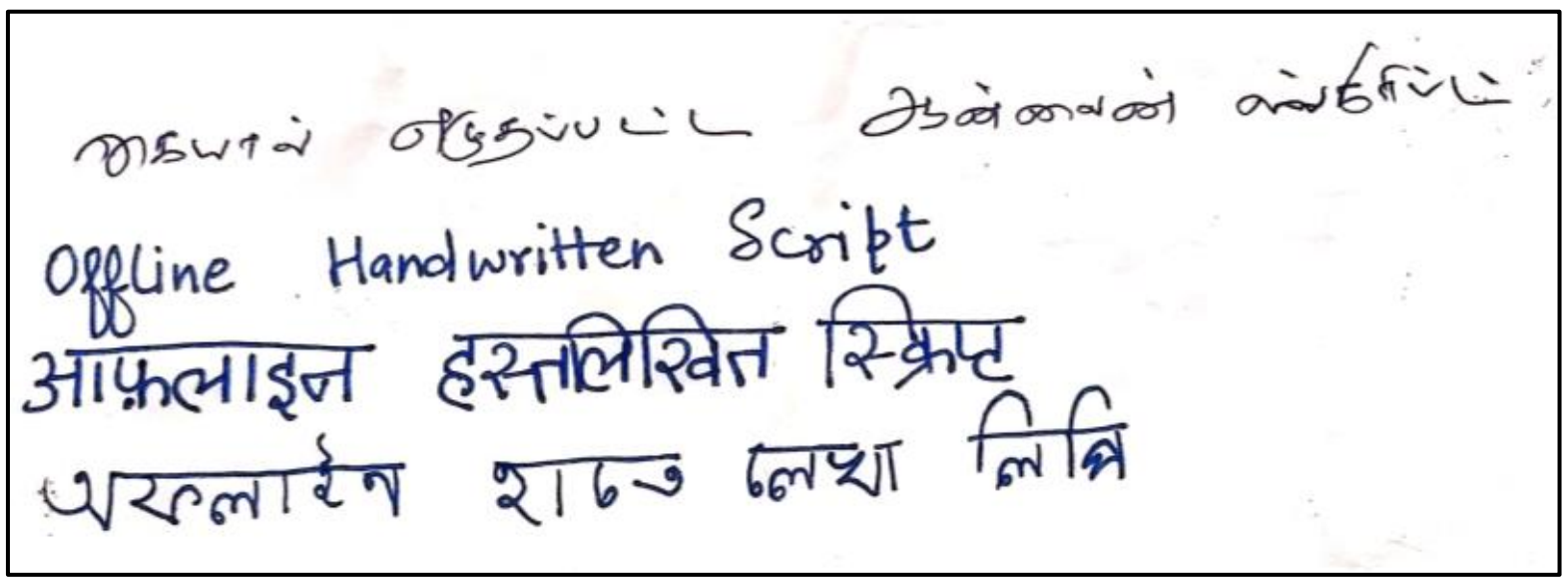

***Figure 5.1 – Handwritten offline Script***

This double training of LSTMs is very beneficial for overcoming various limitations of Recurrent Neural Networks as BLSTMs can be trained using all information in the past and future of a time step. These modified versions are excellent when it comes to understanding the context, but their proper application has not been found yet.

*4) Feature Extraction:*

Feature extraction is the process of selecting the values from the data which are relevant, informative and non-redundant. Overall leading to better interpretation .it starts from the original data and goes on to choosing the relevant and appropriate data. It is a part of dimension reduction since by feature extraction the unnecessary data is rejected and the size or the dimension of the data is reduced. It reduces the dimension required to describe the data since the non-relevant dimensions are dropped.
Some Feature Extraction techniques are:

1. *Autoencoder*: This is useful for efficient data coding. By learning the coding from the original dataset, the key features in the data are identified.
2. *Thresholding*: This is the simplest method for image segmenting. Here a threshold value is set the values below the threshold values are set some value while other values are given some other value this is useful for creating binary images.

*5) HMM:*

Hidden Markov Models (HMMs) are a class of probability model which allows the prediction of a sequence of unknown (hidden) variables from a set of observed variables. It states that the future is independent of the past given the present values. That is with the knowledge of the present state the probability of the future states can be calculated.

## II. TYPES OF SCRIPTS

In general, there are two ways of script acquisition online and offline [4,26]. Offline scripts are generated using scanned images like office documents. Offline scripts can be generated using scanned images from scanner, mobile phone pictures, input bitmap images. On the other hand, online, handwritten scripts are a representation of vector strokes. They are collected for digital tablets by the movement pen tips on the digitalized touchscreen; mode of input can be trackpads, mouse pointer, touch screen, stylus, etc.

### *A. Online Script Recognition*

Online scripts [4-6] are written using a stylus or pen on the digital platform directly like tablets. Online handwritten scripts are represented as

i. Vector for strokes
ii. Time series of coordinates

In the online handwritten text, the building blocks are basic strokes. Strokes are a collection of pixels points form one point to another; they are calculated using the path traveled from the first pen down to the last pen up position. Basic strokes can represent a whole character or apart. A single character can be represented by using a single stroke or a set of multiple strokes, but a single stroke cannot represent more than a character.
Recent trends show that nowadays more research is inclined towards the identification and classification of online scripts. Online Handwritten text identification is a known problem in machine-learning society. Latest Machine Learning methodologies like Deep learning and time series like LSTM [4,6] models are used for character recognition.

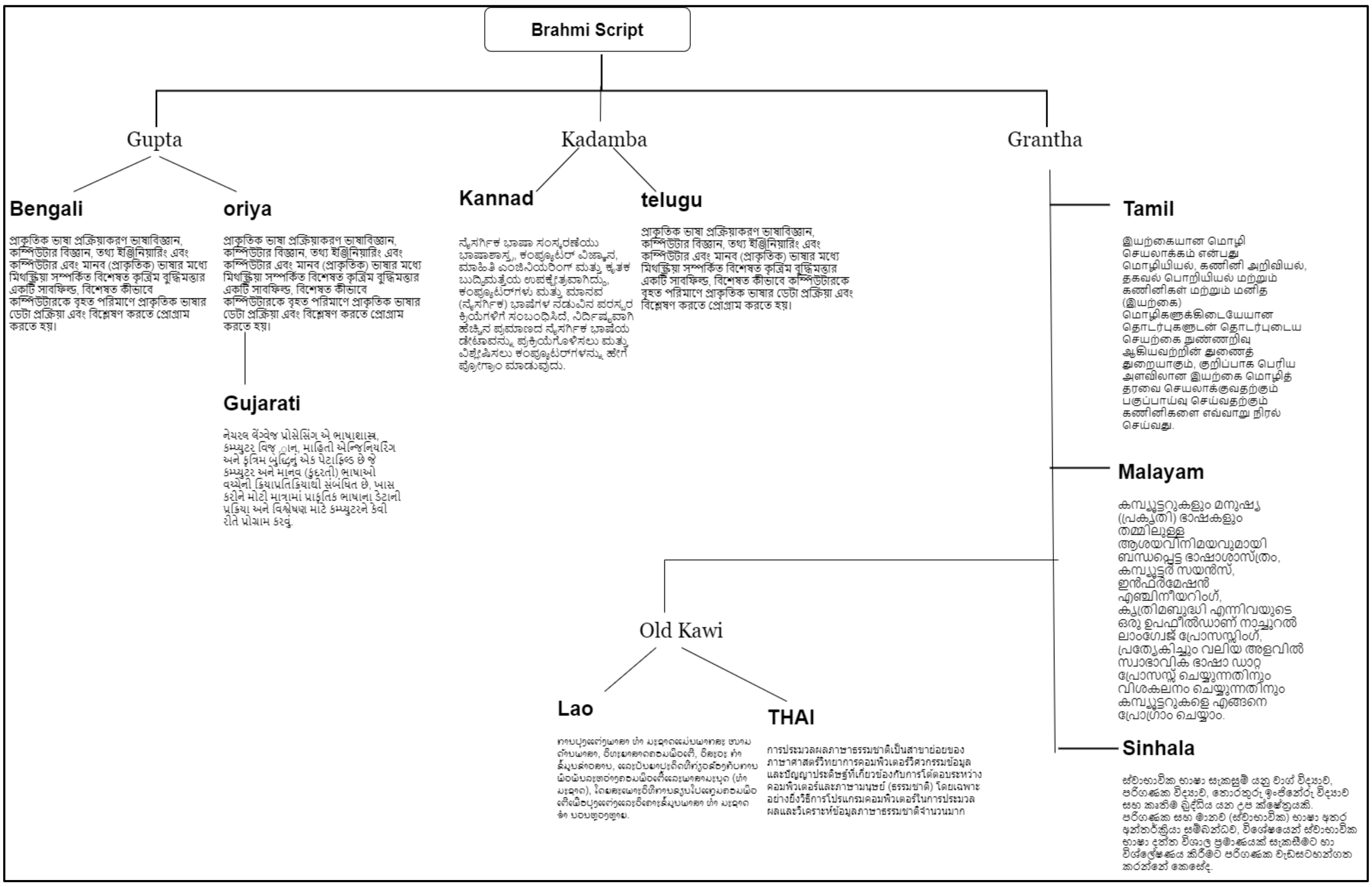

*Figure 6 – Classification of Indic Scripts*

### B. *Offline Script Recognition*

Offline handwritten Scripts [10] are anything written on physical paper using a pen or paper. They are captured using the digital scanner or camera lenses. They are represented as dots of pixels as images instead of vectors as in online scripts. Offline Scripts Recognition is the most common and the oldest problem in Machine Learning and Artificial Intelligence Society. They are harder to recognize and take more processing resources as compared to online script recognition. Before Script identification a lot of image pre-processing needs to be done. Advance machine learning and Deep Learning technique are required for its processing like Convolutional Neural Network.

## III. PROPERTIES OF INDIC SCRIPTS

Scripts are used for symbolizing the writing systems used in various languages. In India, there are 12 major scripts which are used:

- i Devanagari
- ii Bangla
- iii Gujarati
- iv Tamil
- v Kannada
- vi Telugu
- vii Malayalam
- viii Manipuri
- ix Sinhala
- x Gurumukhi
- xi Urdu
- xii Oriya

Out of these scripts, Urdu is the only script that is written from right to left as it is derived from Persian Script. The remaining scripts are written from left to right and are believed to be derived from the Brahmi Script prevalent in 300 BC. These 11 scripts are known as Indic Scripts.[45]

Most of the Indic Scripts except for Tamil and Gurumukhi, contain compound characters formed by the combination of two or more basic characters (vowels and consonants). These compound characters have a more complex structure than the basic characters. [43,44] In some Indic Languages the shape of a character depending on how the consonants and vowels are placed concerning each other. These characters are known as modified characters. In Indic languages, there are more than 300 characters [42].

Most of the text line is partitioned into three designated zones. The topmost denotes the part above the headline, the middle zone contains the structure of the basic character and the last zone contains the main character body.

It is a very common situation that while writing information such as scientific data or technical information the writer might use words or letters from English script. These scripts are known as mixed script documents.

## IV. PREPROCESSING

Several document analysis operations are required to be performed before the text in the scanned document can be recognized [46]. Some of these common operations are listed below:

- i. Thresholding
- ii. Converting a grayscale image into a binary Black-white image
- iii. Noise Removal
- iv. Line Segmentation
- v. Word and character segmentation

### *A. Thresholding*

The main purpose behind thresholding is to extract Black ink from a white background. The histogram representing the gray-scale values of a document image has both High peak (indicates white background) and low peak (indicating the black ink). The threshold gray-scale value is an optimal value between the two peaks.[46]

### *B. Noise Removal*

Digital scanning of images can introduce noise due to scanning devices or medium used for transmission. Multiple smoothing operations such as selective and adaptive stroke filling are used to remove this noise and preserve the connectivity of strokes.[46]

### *C. Line Segmentation*

Line Segmentation is the process of segmenting the handwritten text into lines. A simple approach for line segmentation is to study the horizontal profile of histogram at small skew angles. This approach is however very difficult to implement for handwritten text.

### *D. Word and Character Segmentation*

Word Segmentation is based on the notion of identifying the physical gaps between the words in a line. However, this approach does not yield much success in handwritten text recognition due to the different writing styles of the writer.

## V. METHODS FOR EVALUATION FOR OFFLINE SCRIPTS

### *A. Recognition of visually similar characters*

#### *1) Overview*

One of the most widely used methods for automatically converting digital document images to machine-encoded texts is OCR. A variety of OCR techniques have been developed in the past. An important and incredibly challenging issue in handwritten character recognition is differentiating between visually similar characters in Asian Languages.

The most important phase/step of the OCR process is the character recognition phase where the image is transformed into text. It is the phase where existing methods were unable to provide satisfactory results, particularly in the area which involves recognition of visually similar characters found in several Asian alphabets. This confusion which arises during the recognition of similar characters is a very challenging issue in OCR. The authors [2] proposed a solution for the same.

#### *2) Preprocessing*

Various preprocessing techniques such as Thresholding, noise removal, line segmentation, and word and character segmentation as discussed in the previous sections are employed to prepare the document for analysis

#### *3) Determining the discriminative regions*

All visually similar characters have minute discriminative areas that help in differentiating between them. The identification of discriminative regions is achieved with the help of feature selection techniques such as Fussed lasso

#### *4) Feature Selection - Fussed Lasso Technique*

The Fused Lasso is a regression analysis method that performs both features selection and regularization. This is an extended version of the original Lasso Technique

#### *5) Feature Vector Generation -locality preserving projections*

Locality Preserving Projections: These projections are used for reducing the dimensional space of the feature vectors to reduce space complexity.

#### *6) Optical Character Recognition(OCR)*

Optical Character recognition is achieved using a standard classifier for classifying character images using the resulting feature vectors from the feature extraction phase. The proposed method for OCR resulted in high accuracy as compared to traditional approaches.

#### *7) Advantages of the proposed solution*

- i. The proposed method [2] outperformed traditional/existing character recognition models by huge margins
- ii. The test accuracy for Thai language characters was found to be the highest.
- iii. The algorithm has great potential for similar character recognition is Indic languages

#### *8) Limitations*

- i. Accuracy of the model may not be as good for other languages as for Asian Languages
- ii. Constraints on the performance of HOG (Histogram of Oriented Gradients) feature descriptors

### *B. Using Fully Convolution Neural Network(FCN) for recognizing handwritten text*

#### *1) Overview*

Even though today's generation is shifting towards a paperless world, there are still several scenarios in which handwritten communications is preferred. For example, a traditional pen and paper or modern stylus and tablet. Handwritten text is more convenient, resourceful and efficient. Numerous documents such as paying invoices and question papers often consist of fill-in-the-blank fields that need to be converted to digital form.

This method [1] introduces an offline handwriting recognition algorithm which utilizes the features of a standard convolutional neural network to recognize the stream of word blocks. The proposed method resulted in precise symbol recognition and was able to work with a very large dataset. The proposed algorithm makes use of a lexicon for symbol prediction and recognition.

*2) Preprocessing*

As explained in previous sections preprocessing techniques such as deslanting and normalization are widely used, the authors [1] uses a more cost-effective solution to achieve the same. The method used is similar to the sliding window approach which does not require the use of multiple scanning. Furthermore, the proposed method does not require pre-defined lexicons and symbol boundaries for efficient cleanup.

*3) Word Segmentation*

identification of common English words (the, is, am, are, a, and etc.) is done by using an optional Lexicon CNN. If no common words are found a Block Length CNN is used

*4) Optical Character Recognition(OCR)*

Character Recognition is done using FCN called symbol prediction FCN. The authors [1] proposed the following techniques:

i. Alignment of Symbols: The input word block stream is resized to 32 X 128 and then given to a Block CNN which calculates the total number of symbols N.
ii. Prediction of given Symbols: Symbol prediction is done with the help of Symbol Prediction FCN. The used FCN model has the same characteristics as Block Length CNN, but symbols are used as classification labels.
iii. Even Filter Alignment: This method aligns the output of Symbol Prediction FCN with the ground truth symbols to increase the performance of the proposed method
iv. Receptive Field Filtration: The symbols used in the input word block stream may not be of equal width. This technique is used for making each symbol of equal width

*5) Advantages of Proposed Solution*

i. Same or Better results than traditional pre-processing Techniques such as Deslanting and contrast normalization
ii. This method can recognize common words as well as infinite symbol blocks
iii. Robust to Large Symbol sets

*6) Limitations*

i. The complexity involved in finding the lexicon size
ii. Gradient Propagation issues

*C. Benchmarking on offline Handwritten Tamil Character Recognition using convolutional neural networks*

*1) Overview:*

In this work, Benchmarking on offline Handwritten Tamil Character Recognition using convolutional neural networks, a CNN model has been used to identify and classify both online as well as offline Tamil character, the features have been extracted automatically using the convolutional layer. The dataset used in this model was provided by the HP Labs India which consists of handwritten Tamil characters on HP tablets. The Model's Architecture consists of multiple Convolutional layers, Max-Pooling layer and fully connected layers as well.

*2) Pre-processing:*

In the proposed method [7], the dataset consisted of over 80,000 scanned images of different Tamil characters. Also, different image formats were present too. And the dimensions, as well as the coloring, were different too. For this, the images were first resized to the same size using a bilinear interpolation technique. bilinear interpolation is a resampling technique popularly used in image processing. After resizing the images, they are all converted to the same color encoding which is grayscale with the scale range being 0 to 1.[7]

*3) OCR:*

The proposed model presented in the paper Benchmarking on offline Handwritten Tamil Character Recognition using convolutional neural networks, consists of 9 layers, there are 5 Convolutional layers,2 max-pooling, and 2 fully connected layers. Also, the activation function used here is ReLU which is a nonlinear activation function, and the input images of 64x64 and the output layer are a SoftMax classifier with 156 classes.[7]

*4) Parameter:*

Different parameters and their corresponding values were taken into consideration and the best-suited values were chosen using experimentation and analysis. The activation function used is ReLU which is the most suitable activation function used in image related neural networks. To maintain the image size and dimension no padding is added and the stride value is 1 also different batch sizes were taken into account but due to memory issues and performance factor 64 was chosen. The model was trained for 100 epochs and the after each epoch the validation accuracy was also checked.[7]

*5) Result:*

After which the testing accuracy was evaluated from the model. While the training accuracy was initially very high and testing was low indicating overfitting. A dropout rate was added to the model which gave a training accuracy of 95.16% with validation score of 92.74% and the testing accuracy of 97.7%.

*6) Merits:*

The model made in this paper had a training accuracy of 95.16% while the testing accuracy was 97.7%. also, the dataset was very large and highly detailed with many features and classes which can be useful for further training this model or for training other models as well.

*7) Demerits:*

Though the training and testing accuracy was very high, yet the model fails to differentiate between few characters which were very similar to each other. Thereby model classified a few characters wrongly.

### *D. Devanagari character recognition in printed and handwritten documents using SVM*

*1) OVERVIEW:*

In this paper, Devanagari handwritten and printed text recognition using SVM, the model proposed is trained using the scanned version of printed documents collected from different government websites. The model is trained on Devanagari script and comprises of 3 languages Hindi, Sanskrit and Marathi. The model works by first removing the Shiro Rekha of each character and then identifies and classifies each Shiro Rekha less character using an SVM classifier. The images are first pre-processed and then features are extracted and then finally the model is created from the obtained data.

*2) Pre-processing:*

In the paper, Devanagari handwritten and printed text recognition using SVM the proposed classifier model used is SVM which has been designed such that it accepts scanned images of printed or handwritten Hindi, Sanskrit, and Marathi text. After that image pre-processing is performed on those images, the pre-processing used in model involves removing the noise from each of the images, after that converting the images to grayscale from RGB color coding, and then the images are normalized with skews of each images being corrected. The variable and the path locations are also set up in this stage.[8]

*3) Segmentation:*

After the pre-processing is performed on the images, they are passed further for the feature extraction, the segmentation is done in the paper Devanagari handwritten and printed text recognition using SVM [8]. the primary step in the segmentation process is the extraction of the horizontal line which is called Shiro Rekha. This line is first detected by comparing the black pixel density in the image and the highest density being in the black line. After the line is detected. The value of every pixel is made equal to zero this is done so the line can be completely removed and only the necessary characters are remaining. These characters are known as Shiro Rekha fewer characters. These characters are then passed further for feature extraction.

*4) Feature Extraction and classification:*

After the segmentation is done in the previous stage and the Shiro Rekha is removed, all the Shiro Rekha fewer characters are obtained. The features of these characters are extracted. This is done by representing them in form of 2-dimension matrix (rows x Columns) where rows and columns are equal, by the order of the magnitude of the black pixels in the matrix the relevant features are extracted from the images, these features are then used further to build the SVM model. For the classification, the features obtained from the image are matched from the classes which are have been pre-defined.

*5) Dataset:*

The dataset used in the paper Devanagari handwritten and printed text recognition using SVM consists of the scanned documents which have been collected from different government sites and the handwritten document. Documents are in all the languages mentioned and then divided for training and testing.

*6) Merits:*

The model proposed in this paper was able to work very well for the Devanagari script and had a very diverse dataset consisting of 3 languages Sanskrit, Hindi, and Marathi .and had a staggering accuracy of 99.23% for Hindi documents, 97.23 % for Sanskrit documents and 98.61 % for Marathi documents

*7) Demerits:*

The Model was more efficient for Hindi and Marathi language more than the Sanskrit language, this was because a few Sanskrit characters were misclassified. The main reason for this misclassification was because Sanskrit characters have a similar geometric structure and their complex conjuncts which caused ambiguity for the model thus causing misclassification.

### *E. Devanagari numeral recognition using CNN and genetic algorithm.*

*1) Overview:*

This work proposes a hybrid of both CNN and Genetic Algorithm for handwritten character recognition. The proposed work first pre-processes the image dataset and then the features are extracted using the sparse autoencoder. After which the CNN is applied on the images and the output of the CNN is used as the input of the genetic algorithm which gives us the best candidate for the model and then passes it further to Fully Connected layer and then finally to L-BFGS which improves the model by checking and evaluating the suitable weights for the fully connected layer.

*2) Pre-processing:*

In the paper Devanagari numeral recognition using CNN and genetic algorithm [9], the pre-processing first involves the images being converted from RGB format to grayscale format. This conversion is because different images have different color encoding and conversion to grayscale removes ambiguity in the color-coding. After all the images are converted to grayscale the pixels are normalized and a threshold is applied to each pixel. After this, all the pixels are normalized and then rescaled and then all images are resized to the same size which is 28x28 and a padding of 2 pixels is done.[9]

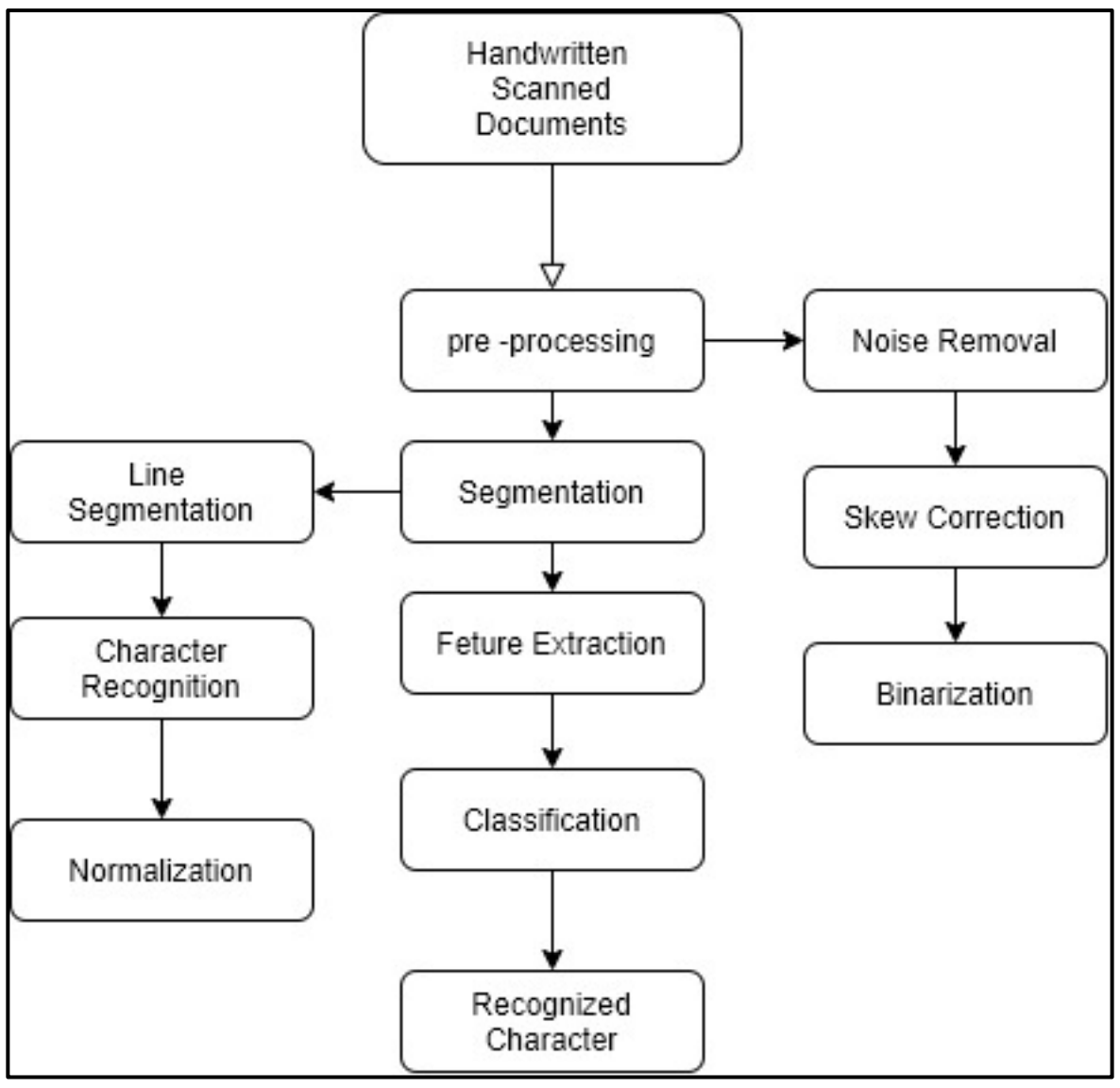

***Figure 11 – Flow diagram for offline character recognition***

*3) Feature Extraction/Image segmentation:*

The sparse autoencoder is used for getting the most relevant and necessary features from the dataset, autoencoder is a deep learning model which learns approximation from input data. The sparse encoder also reduces the resources needed to present the data thereby reducing the dimension. The number of nodes in the CNN layers are also reduced so that the low dimension features are extracted. The sparse encoder is trained using hyperparameters and L-BFGS gradient optimization [9].

*4) OCR:*

In this paper they have implemented using the CNN and the output of the that is passed as the input of the genetic algorithm

*5) CNN Architecture:*

The Convolutional neural network used in this paper, Devanagari numeral recognition using CNN and genetic algorithm [9] comprises an input layer with dimension 32*32*3 as the input which represents the dimension of each of the images. After the input layer, a convolutional layer is applied with the filter size 9*9 and has 256 features. After this layer the parameters in the model are increased, for this Mean Pooling layer is implemented with the kernel size 8x8. With the pooling layer, the number of the parameters is reduced thereby decreasing the model complexity and further improving the efficiency. After the implementation of pooling layers, the obtained dimension in the output is passed further for the genetic algorithm input. The output of the CNN model acts as the input of a genetic algorithm.[9]

*6) Genetic algorithm implementing:*

In this paper Devanagari numeral recognition using CNN and genetic algorithm, the genetic algorithm is implemented by first generating a set of candidates. the output obtained from the CNN model is passes into candidates as their respective lengths. After this, the chromosomes are generated. Next, a few candidates are selected at random and the fitness of the candidates is calculated. Amongst them, the candidate with lower fitness value is selected since the one with higher value would mean a longer chromosome thereby making the model more complex. This process is repeated until all the best candidates are chosen. After the candidates are chosen the weights encoded in them are mutated by a small value and this is carried out for 100 or more times. The candidates after this mutation are passed into the final layer and these candidates are acting as the SoftMax layer's initial weight.

*7) Merits:*

The model proposed here had very high accuracy for training as well as testing. But since CNN requires a huge number of iterations, it is very expensive to train a model. This genetic algorithm is also imparted in the model which helps in initializing the weights of output layers.

*8) Demerits:*

Since the number of iterations was over 500 in the initial phase, there were chances of the model being overfitted and thereby decreasing the testing accuracy. Also, since the number of iterations was large, the length of the chromosomes became large thereby increasing the model's complexity, which makes the model lesser efficient.

### *F. Offline Character recognition*

*1) Preprocessing*

Preprocessing techniques used in this paper [47] are:

i. Noise Removal: Various types of noises (Gamma noise, Gaussian Noise, Rayleigh Noise) are removed using Filters such as Ideal Filters, Butterworth Filters and Gaussian Filters.

ii. Skew Detection/Correction: Skew Detection and correction is used for proper alignment of paper with the coordinate system of the provided scanner. A variety of skew detection techniques such as projection profiles, connected components, Hough Transform are used for the purpose [47]

iii. Binarization: Conversion to the binary image is achieved with the help of adaptive thresholding, Global Thresholding, and Otsu's Method.[47]

*2) Segmentation*

A series of grouping and spitting algorithms are used for segmentation [47]. Region finding used for

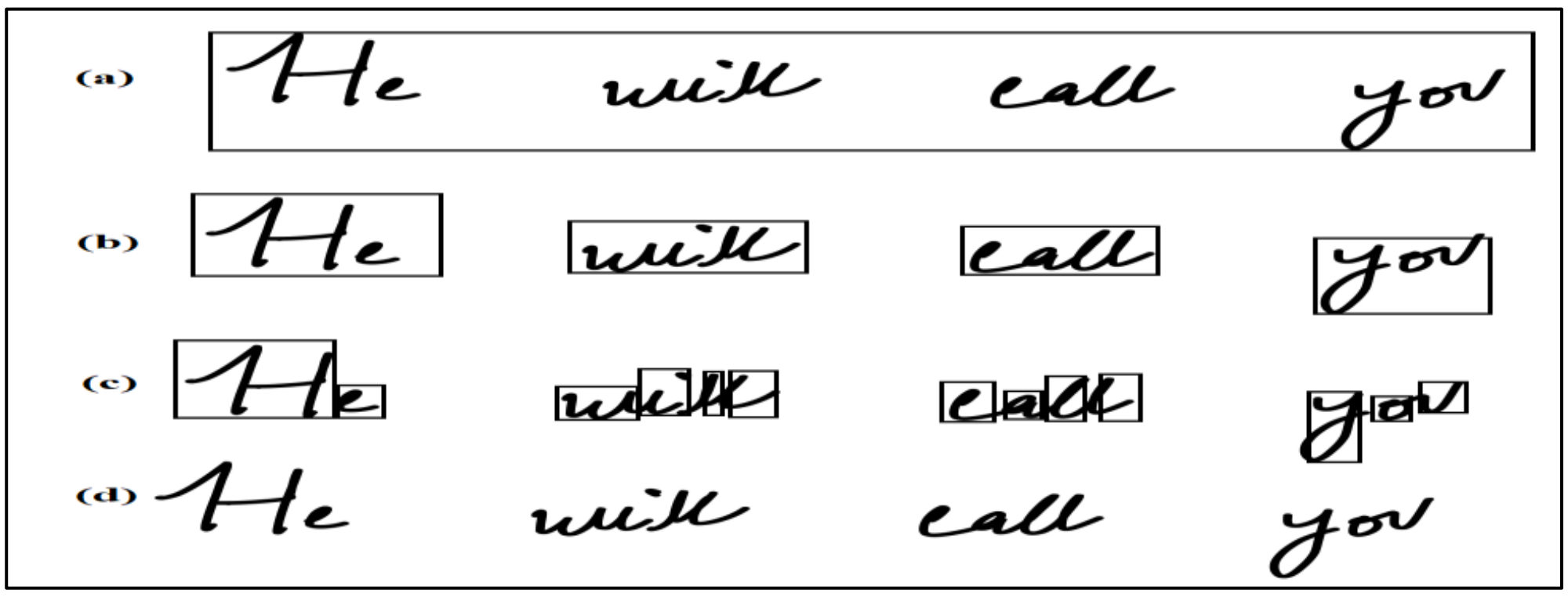

*Figure 12.1 – (a)line segmentation (b)word segmentation (c) character segmentation*

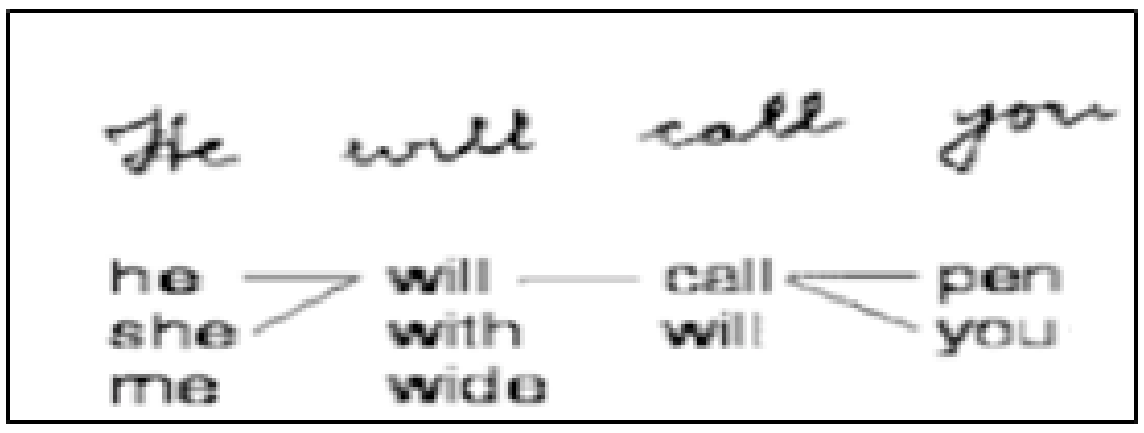

*Figure 12.2 – probabilistic based sentence recognition*

identifying all separate regions. All the pixels are given 'ON' or 'OFF' labels where 'ON" symbolizes data regions. The given image is examined pixel by pixel until a matching 'ON' value is found. As soon as a pixel is found with 'On' label a unique number is assigned to it and its nearest neighbors are searching for 'ON' labels. The process continues if no 'ON' labels are found [47]

*3) Feature Extraction*

The paper [47] handles the identification of the following feature categories:

i. Statistical Features: Various Techniques such as zoning, crossings, and distances, projections are employed to detect statistical features
ii. Global Transformation and Series Expansion: A variety of techniques such as Fourier Transform, wavelets, moments, Gabor Transform are used.
iii. Geometric and topological features: Structural features such as loops, T-point, etc. are used.

### *G. Offline Recognition of handwritten script*

*1) Preprocessing*

Various preprocessing operations are required for document analysis before the text can be recognized. The given paper [46] has performed preprocessing operations in the following manner:

Thresholding: Thresholding is performed by identifying the optimal value between the high and low peaks of the gray-scale histogram of the document image.

Noise Removal: Noise is removed from the scanned images using various smoothing operations like selective and adaptive stroke filling.

Line Segmentation: A clustering technique is employed to form a collection of minima of various components to identify the different handwritten lines in the document image.

Word and Character Segmentation: Various writing styles are studied and converted into a function for word and character segmentation.

(a)Line Segmentation (b) Word Segmentation (c) Character Segmentation

*2) Optical Character Recognition*

The authors use the following algorithm for OCR [46]:

i. Provided a handwritten sample, the first step is to perform character segmentation
ii. The next step is to extract the micro-features for each character
iii. Each handwritten sample is characterized by the number of micro-feature vectors that correspond to characters in the given sample.

*3) Word Recognition*

The authors use a combination of analytical and holistic approaches to achieve high performance for word recognition algorithm [46]

## VI. Methods for Evaluation for Online Scripts

### *A. Word recognition for online handwritten based on RNN models for Bengali and Devanagari scripts that use horizontal zoning method*

*1) Overview*

This work [4] proposes an RNN model variants (LSTM and BLSTM) model for online handwritten word recognition for both Bengali and Devanagari scripts. This work is proposed for the first time in and is the novelty of this method Each word is first divided into 3 horizontal zones before training is carried out for the basic strokes using LSTM and BLSTM variants of RNN model. To show a comparative performance analysis recognition is also carried out using conventionally used HMM models for the recognition system. Results demonstrate that the proposed work using zone segmentation and RNN based LSTM–BLSTM learning models outperform the existing conventional recognition systems which includes HMM models for both the scripts. The drawbacks of the conventional HMM-based model are overcome by the proposed system [4]. The conclusion can be derived that the proposed method can provide new insights into developing similar systems for other Indian scripts.

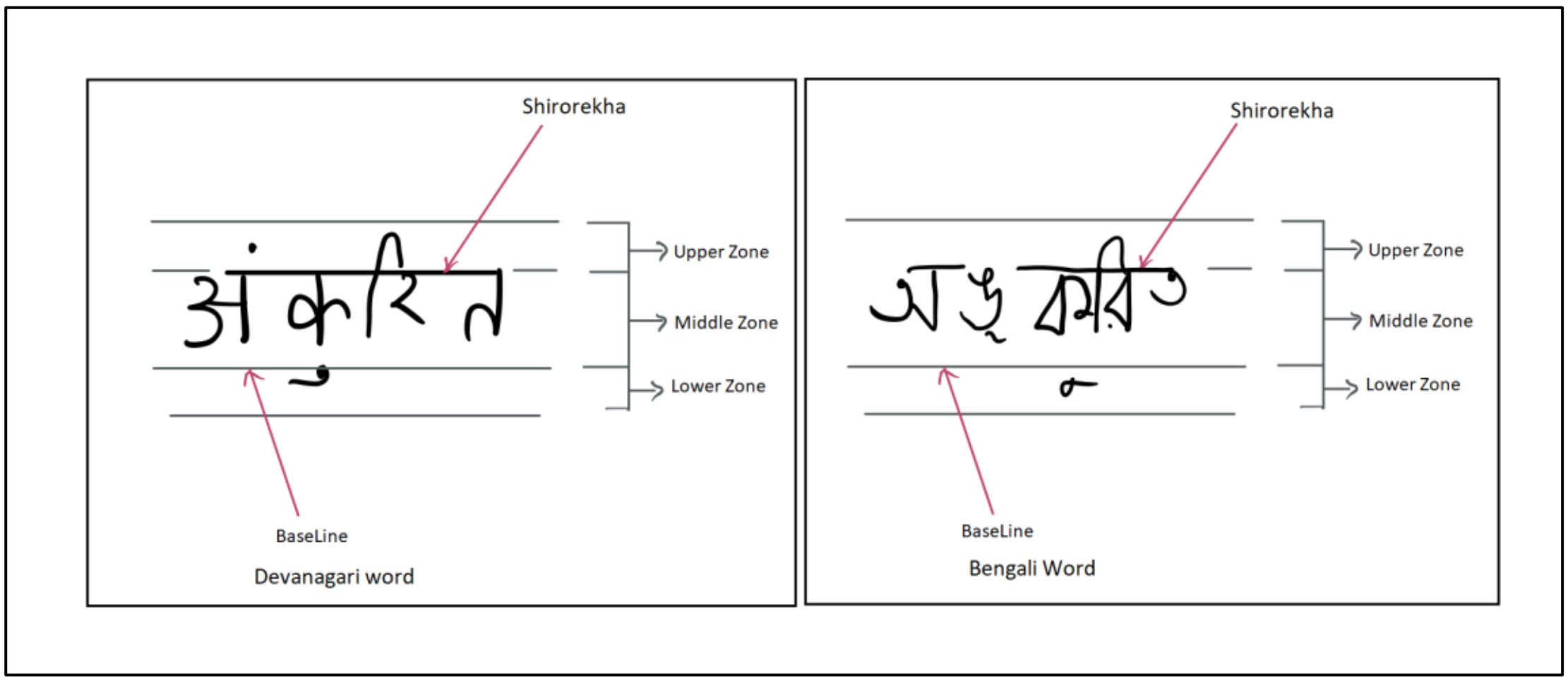

***Figure 13.3 – Zone Segmentation***

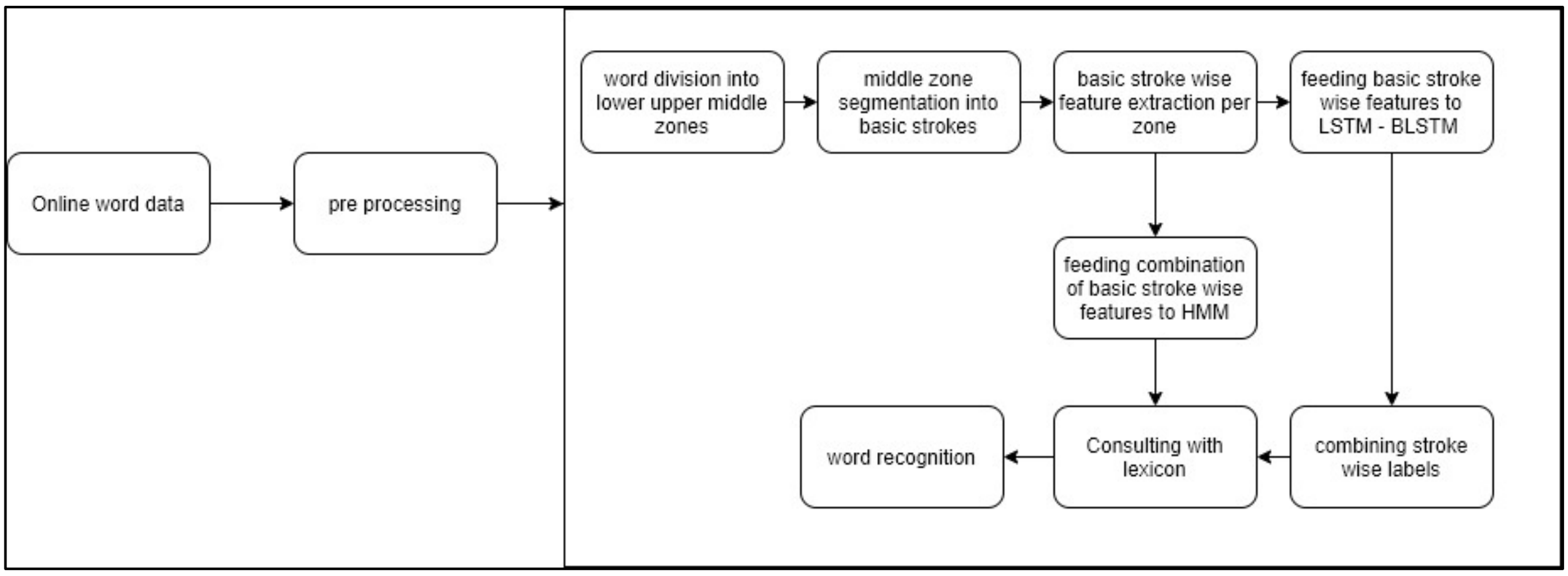

***Figure 13.2 – Flow diagram for RNN models for Bengali and Devanagari scripts that use horizontal zoning method***

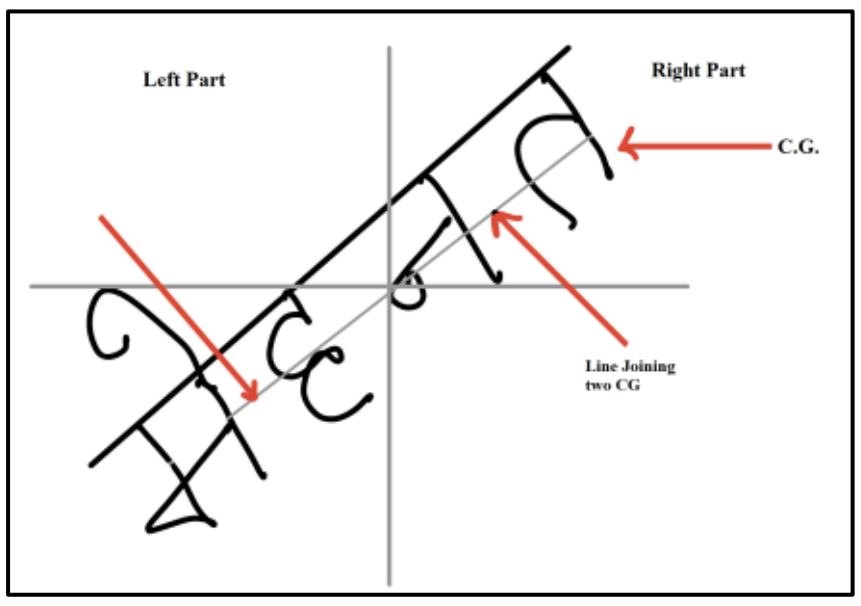

***Figure 13.1 – Skew Correction***

*2) Pre-processing*

The authors [4] discussed greatly the problems and the ambiguity while writing the same word or phrase by different persons. The authors also proposed advance statistical and vector-based approaches for solving the same. When collecting data online due to the variation of speed in writing some points remain close to each other (when writing slow) and others might be far apart when writing fast. This leads to the variation in the number of strokes. The strokes may differ based on the style of writing between person to person. These problems are tackled by pre-processing of the time series of the word

The methods for pre-processing discussed in this methodology are

i. *Interpolation* – This method is applied to the words and the strokes to counter differences in the speed of writing the word.

**ii.** *Smoothing* – due to trembling in some people the path of the basic stokes might vary from the actual

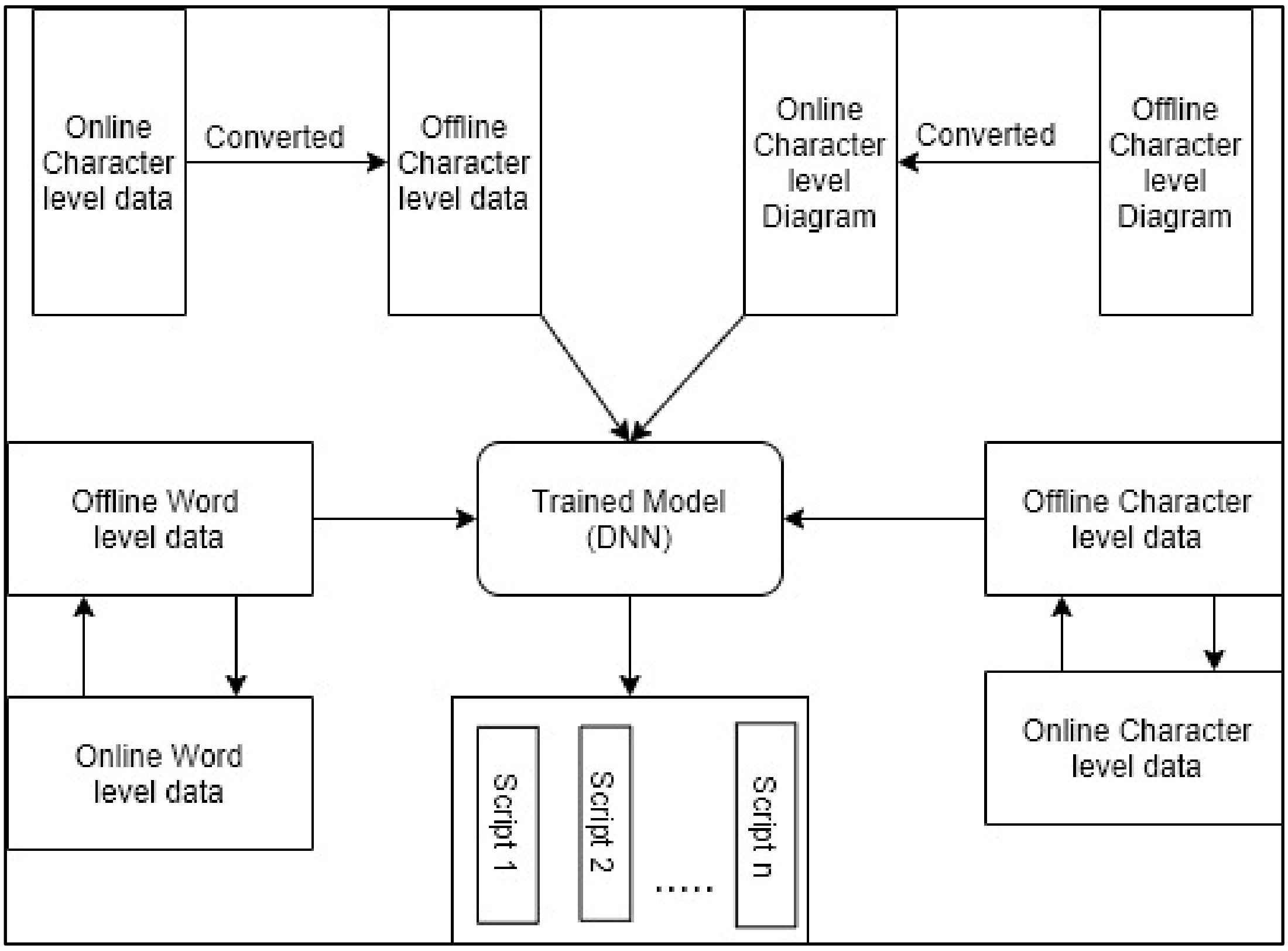

***Figure 15 – Modality conversion and training using CRNN models***

stoke intended. This can be solved by applying smoothing filters

**iii.** *Re-sampling, Normalization* – the height of the same basic strokes might differ for the same strokes this could be solved by Resampling and Normalization

**iv.** *Skew correction* – In many scripts the word is written in skewed fashion i.e. the word differs from the original x-axis. This can be solved by Skew correction

*3) Zone segmentation (the novelty of the paper)*

In Indian scripts, the words can be divided into zones based on (Shiro Rekha for Devanagari and matra for Bengali). Zone segmentation is proposed in this method to reduce the possible number of combinations of basic stroke wise labeling (RNN) or characterized labeling for HMM for a word. This reduces the total possible combination that would be required to feed into the network or the HMM model this increases the overall efficiency and accuracy.

The authors propose a novel approach to segment each word in three zones based on Matras and Shiro Rekha for Bengali and Devanagari scripts. These zones are classified as upper-middle and lower zones. Upper and Lower containing the vowels or the Matras and the middle containing the consonants. This method reduces the overall possible combination need to feed the network, in turn, increasing the accuracy.

Middle and the lower layers are separated by detecting the horizontal line present in Indic scripts. These modifiers are classified separately and detected using a different network

*4) OCR*

The authors [4] proposes Two models for letter detection and word identification 1st being LSTM and BLSTM models and the other being HMM implementation. These two models are used to give a comparative analysis of both models.

They created two models of RNN LSTM (long-term short-term memory) and BLSTM (Bidirectional long-term short-term memory) for the identification of scripts characters. These are fed a sequence of time-series data after pre-processing and segmentation. The HMM model is fed with each character which is segmented after pre-processing. The HMM model predicts the letter with maximum possibility.

*5) Merits*

The proposed work can identify basic strokes using Models like BLSTM even after stroke segmentation which cannot be produced by using standard models like HMM.

This system overcomes the limitation of HMM which cannot carry information about the past state as RNN contains multivariate internal states which contain information about wholes series of

*6) Limitations*

One of the major limitations of this system is the identification of similar character appearing in the middle segment of the word. This led to most of the occurring errors

*B. Multi-Component sinusoidal algorithm for representing online handwriting scripts*

*1) Overview*

The authors [5] propose various methods of extraction of temporal and spatial attributes for individualization and analysis of handwritten patterns. It proposes a multi-component sinusoidal model [5,6] that can be used to represent online handwriting. Method reforming online-based signatures and to develop an online handwriting character and word recognition model is proposed in this work. The works also explore a combination of currently existing point-based characteristics and sinusoidal parameters in the handwriting recognition system. Each word in an online script can be represented as a set of vertical and horizontal velocities in time domain thus creating a pre-processing method for time series prediction of the word. This work [5] is used for the conversion and pre-processing of online scripts and used by other models and proposed models [6].

*2) Pre-processing*

Before conversion to sinusoidal form pre-processing of the word and data is required to remove any ambiguity present in the script. This prevents error in conversion which may lead to higher error rate in identification

As discussed in previous segments [4] this method uses standard pre-processing methods/techniques such as interpolation, Smoothing, Re-sampling, Normalization, skew correction. These methods are greatly discussed in paper [4]

*3) Feature Extraction*

Online scripts contain strokes and stokes are a set of time-based vectors. Strokes have features and these features can be examined and analyzed to perform analysis and recognition. These features like velocities and densities can be extracted from time-based vector online scripts.

This work discusses two major feature extraction methodologies

*Sinusoidal feature extraction* – In this method of feature extraction for representation no. of components (K) is required, this is calculated/analyzed by reconstructing SNRs of signals. The dimension of the vector is calculated by the analyzed value K. In the proposed work dimension calculated is 3 therefore 3 components are extracted as parameters. Only frequency and amplitude parameters are used as a featuring vector.

The formula used for the sinusoidal feature set

$$\mathbf{F}(\mathbf{t}) = [\boldsymbol{a}_x^1(\mathbf{t})\boldsymbol{a}_y^1(\mathbf{t})\boldsymbol{\omega}_x^1(\mathbf{t})\boldsymbol{\omega}_y^1(\mathbf{t})\ldots\ldots\boldsymbol{a}_x^K(\mathbf{t})\boldsymbol{a}_y^K(\mathbf{t})\boldsymbol{\omega}_x^K(\mathbf{t})\boldsymbol{\omega}_y^K(\mathbf{t})]$$

*Point-Based feature extraction* – To obtain this feature two-dimensional analysis along the x-y plane is performed. Features are attained at every point [5] $(x_t, y_t)$ which is between points $(x_{t-r}, y_{t-r})$ and $(x_{t+r}, y_{t+r})$. here r represents a very small value. This method calculates the feature for a point using statistical and graphical methods by noting the change that occurred for the length r.

Features calculated using point-based analysis are (with number of dimensions)

- The derivative of x and y coordinates – 2D
- Aspect ratio – 1D
- Curliness at any position – 1D
- Writing direction – 2D
- linearity calculated using an average of squared distance – 1D
- the curvature at each point – 2D
- trajectory slope at any given point – 2D

*4) OCR*

*Word Recognition*

For this algorithm used IRONOFF online handwriting word dataset/database for analysis of the performance of proposed work. Multiple model variants of the Hidden Markov Model like GMM-HMM and DNN-HMM have been used for the analysis of performance.

The DNN-HMM consisted of 2 hidden layers with 300 neuros each. The input parameter was a 3D set of features. The model received an input of 22-dimensional features which are calculated using sinusoidal features for K=2

The GMM-HMM model is a point-based feature model consisting of a 14-dimensional set of features.

*Character Recognition*

The proposed character recognition model is created for GMM- HMM. The HMM model is trained by using a left to right linear topology. The Gaussian mixture model is used to calculate observation probabilities. The HMM parameters (number of states and mixtures) are optimized and set between the range 10 and 25. This work also uses a combined DNN-HMM model to demonstrate the efficiency of the proposed features model.

*5) Merits*

The primary advantage of this method is that it proposed a method to represent the handwritten word in the form of a feature of sin waves. This can theoretically be inaccurate and would take less time to compute when compared to using raw images.

*6) Limitation*

Major drawback about this method was the length and degree of the wave could not be determined for handwriting and model must be made which could set the degree of the wave which would tune into the degree of handwriting

## VII. New Approaches which incorporates both Online and Offline Scripts

### *A. Handwritten script identification using both offline-online multi-modal deep network for Indic Scripts[6]*

*1) Overview*

The proposed work designed a multiple script identification system that would work as a single modality. The challenge encountered in such work is a handwritten text that has been plagued by inherent provocations because of the freely flowing characteristics of the manuscript, contrasting the standard machine-created text that has a standardized structure throughout. The variation in the style of writing among different entities and complicated. Contours of characters are just some of the principal limitations for handwritten script identification.

Present literature and research for script identification/ recognition are mostly based on removing linguistic and statistical-based methods at the line or word level. Contrast suggested framework relies on the hypothesis that distinctive features are alone contained in the character set of a script.

In the proposed work a multi-modal deep neural network accounts for both online and offline modality of the data as input for script identifying the task so that information from both the modalities can be explored. Handwritten data in both modalities are taken as input or the reverse modality is created through intramodality conversion. Followed by conversion the deep network is feed with the pair of offline-online modality. The proposed model framework alleviates the requirement of designing more than one script identification module on the individual modality. It takes advantage of utilizing derived information from multiple modalities thus making it feasible to work for both offline and online script identification

*2) conversion*

Inter-modality - One of the key steps in the proposed [6] framework is Intermodality conversion. As the proposed work used deep neural network which holds the online-offline pair as their input, a conversion of modality is needed to achieve differing modality from the input data.

*Offline-Online* - The thinning process is used to extract the skeleton image of the handwritten text is which is in turn used to extract a parametric curve. Translations are staged until the centroids of the two turn out to be aligned so that a comparison can be performed between a dynamic exemplar and static image skeleton. After this process, it is converted into a static image. In the proposed work [6] the static image skeleton and dynamic exemplar have been thickened to a width of five pixels. Following the process, two images are matched tracked by the trajectory extraction. Hidden Markov Model has been applied for ascertaining the pen trajectory from a normalized and static image.

*Online to offline* - In Online data the flow of writing is represented by consecutive co-ordinate points. To derive offline data, from the online equivalent. An empty image matrix is defined based on the maximum and minimum difference of (x, y) coordinate values. Thereafter, pixel positions of the empty image matrix are marked basis join consecutive points and online coordinate points serially. The skeleton image of the handwritten data is generated. Morphological filters and thickening operations are performed after this to make the comparable offline word image like the real offline word image data.

*3) Network Architectures*

*Online modality* - In this architecture, the sequence of x-u points is used for the representation of the flow of writing. As the data point is sequential time-based analysis can be done by directly feeding them to the LSTM model and getting the output. But the authors propose a better approach which uses a CLSTM and hybrid or ensemble model of both CNN and RNN. The convolutional part is added to the network to make it more reliable for detecting the properties of scale, shift and distortion invariance. As this uses an analysis of two-dimensional image the property on local connectivity in the Convolutional layer of CNN plays a key role to find the correlation among neighboring pixels. Before feeding the RNN model the data (sequential coordinate points) are passed through CNN layers so that the relation among neighboring points can be accounted for. Thus, it can be concluded that using CNN converts the x-y coordinate points into higher dimensional attributes which take multiple characteristic variants of the individual.

*Offline modality* – In online data, the flow of writing can be represented as time-series based analysis vector strokes in the time domain. The data could directly be feed to an LSTM model to extract the time-based features and perform analysis. But in offline scripts data are in the form of 2-dimensional images and need to be converted into a continuous series of vector data before it could be feed to the network. The conversion could directly be done by tracing the path of the data and sliding approaches, but the features extracted by the vector would be limited. The work presented in the papers uses a convolutional neural network to solve the problem. Before the data is fed to the LSTM mode the data is passed through CNN models which extract other essential features form the image for increasing the accuracy of prediction.

*4) Conditional multimodal fusion*

By using the above two explained network architecture online and offline modalities are interchangeable calculated. Time-based features are extracted from both the modalities using both 1 and 2 dimensional CNN. The sequential feature extracted from the CNN modules is feed to RNN based LSTM and BLSTM models. Pure RNN models are not used directly due to its limitation of exploding and vanishing gradients. To overcome the drawback of the RNN LSTM model is used with 3 multiplicative gates and a memory bank

## VIII. DATASETS USED

In the past few decades, Indic script recognition has seen development in terms of both algorithm performance and availability of standardized dataset. Many of the researchers artificially create dataset for performance analysis of their algorithm. They tend to compare their methodology with other using standard datasets.

The availability of standard dataset is much required for research purposes, without which a comparative analysis cannot be performed with similar work. One of the standard online text recognition datasets in English language is

Table 1.1- Datasets of Printed scripts

| Dataset used | Type of Database | Language | Page count | Characters /words |
|---|---|---|---|---|
| Times-of-India Sanjevani vijaya-Karnataka-Amar-Ujala | Un structured | Kannada, English, Hindi | - | - |
| Digital Library of India (DLI)database | Structured | Bangla, Gujrati, Devanagari, Kannada, Gurmukhi, Tamil, Urdu | - | - |
| TOI, Udyavani Navbharat Times | Un structured | Kannada English Hindi | - | - |
| Supplementary Books for different courses | Un structured | Hindi Bangla English | - | - |

Table 1.2- Datasets of Offline scripts

| Dataset used | Type of Database | Language | Page count | Characters /words |
|---|---|---|---|---|
| IAM | Structured | English | 1539 | 115320 |
| RIMES | Structured | English | 12723 | - |
| EMNIST | Structured | English | - | 145600 |
| THI-C68 | Structured | Asian Languages | - | 13130 |
| NECTEC | Structured | Asian Languages | - | 5M+ |
| HP Labs Dataset | Structured | Tamil | - | 77500 |
| V2DMDCHAR | Structured | Tamil | - | 20305 |
| ISIDCHAR | Structured | Hindi Sanskrit Marathi | - | 36172 |
| Indian Statistical Institute (Kolkata) dataset | Structured | Hindi | - | - |

Table 1.3- Datasets of Online scripts

| Dataset used | Type of Database | Language | Page count | Characters /words |
|---|---|---|---|---|
| IRONOFF | Structured | English | - | 21364 |
| UIPEN | Structured | English | - | 1364 |
| PHD Indic | Structured | Bengali Devanagari | - | 11000 |

*IRONOFF [84] UIPEN [85]*, these datasets contain structured data for word and line level scripting. *IRONOFF and UIPEN* dataset are the most used dataset for word recognition or pre-processing. Choudhary H. et al. [5] and *Bhunia A.K. et al. [6]* have given a comparative analysis of their algorithms using these datasets. *IAM RIMES NIST* datasets are the most common datasets for offline character recognition in the English Language. These datasets have Line granularity and provide data in structured composition for training Neural Network models.

There has been a lot of research done for script recognition system in the English language, and the availability of standardized dataset makes the comparison task efficient. This is not the case for Indic scripts, the data pool for Indic character recognition is very limited. Due to these reason researchers prefer to create their own dataset or work on unstructured – unlabeled data. Although there have been a increase in dataset for Indic scripts the label and the size of the data are barely sufficient for conclusive study.

The most commonly available dataset for Indic script is in Devanagari. These datasets include *Indian Statistical Institute (Kolkata) dataset* studied by *Trivedi A. et al. [9], HP Labs Dataset [86] used by Raghupathy K. et al. [7] and ISIDCHAR V2DMDCHAR by Puri S. et al. [8]*. These datasets are offline handwritten or printed datasets for Devanagari scripts. *PHDIndic* is one of the databases which contains ladled character for Indic script. It contains in total of 11 Indic scripts for both structured and unstructured data.

Most of the Online text recognition script are manually constructed or unstructured data has used them to create artificially. This is due to the lack of standardized Indic script for online data. Bhunia *A.K. at el. [6]* propose a method to convert offline data to online modalities which could be studied further and be used to convert structured – unstructured offline Datasets to online features.

## IX. Comparison Table

Based on the above data, it can be inferred that classifier centric approaches which involve using models such as SVM, CNN, and HMM are widely preferred for handwritten text recognition. These models have shown benchmark results when used for character identification and segmentation. The above models are mostly preferred for Offline handwritten text recognition while Online text being a part of time series analysis requires advance models of RNN and HMM. The benchmark comparison presented by the authors were mostly between the conventional HMM model and RNN Variants such as LSTM and BLSTM.

Other models such as K-NN and NN have also shown good accuracies and are also very easy to implement but the lack of databases for training and testing these models is a limiting factor. Moreover, for most practical problems KNN is a bad model due to lack of scalability. These models require large databased and properly labeled data with less distortion or outliers to predict correctly. Thus, these are majorly used for printed texts instead of handwritten document

A comparison of pre-processing techniques has also been presented in this article. Standard methods that have been used by researchers are Erosion and Dilation Morphological operation binarization, Binarization Noise Removal Skew Correction Morphological Operations, Thresholding. Many other states of art pre-processing techniques have been proposed by authors which show defining result. From the results achieved so far, we can analyse that the selection of the classifier, feature extraction techniques as wells as pre-processing techniques need to be proper to attain good accuracy in recognizing the character.

**Table 2.1- Comparative analysis of Online scripts**

| Researcher | Methodology | | | Type of Script | Granularity | Dataset used | Best accuracy/ performance |
|---|---|---|---|---|---|---|---|
| | Pre-processing | Feature | Classifier | | | | |
| Ghosh R. et al. [4] | Interpolation Smoothing Resampling Normalization skew-correction | Zone segmented stroke-based feature | RNN HMM | Devanagari and Bengali | Text line | Collected data | 99.4 |
| Choudhary H. et al. [5] | Interpolation Smoothing Resampling Normalization skew correction | Vector based stroke features | GMM-HMM DNN-HMM | English | Word | IRONOFF UIPEN | 1.23 WER |
| Bhunia A.K. et al. [6] | Inter modality conversion | Vector based stroke features | CLSTM BLSTM LSTM CNN | Bengali and Devanagari | Page | IRONOFF PHDIndic | 94.7 |

**Table 2.2- Comparative analysis of Printed scripts**

| Researcher | Methodology | | | Type of Script | Granularity | Dataset used | Best accuracy/ performance |
|---|---|---|---|---|---|---|---|
| | Pre-processing | Feature | Classifier | | | | |
| A. Kumar et al. [68] | Erosion and Dilation Morphological operation binarization | Stroke based features | Learning algorithm | Kannada, English, Hindi | Word | Times-of-India Sanjevani vijaya-Karnataka-Amar-Ujala | 98.79 |
| P.P. Yeotikar et al. [70] | Removal-of-non-text-regions skew-correction noise removal binarization | Stroke based features | Simple voting technique | Kannada, English, Hindi | Word | Manually constructed | 99 |
| D. Dhanya et al. [71] | Gabor Filter | Spatial spread and directional feature | SVM, NN, K-NN | Tamil, roman | Word | Manually constructed | 96.03 |
| G.D. Joshi et al. [72] | Text Segmentation, Binarization | Texture based features | Hierarchical script classifier | Bangla, Gujrati, Devanagari, Kannada, Gurmukhi, Tamil, Urdu | Text line | Digital Library of India (DLI)database | 97.11 |
| P.K. Aithal et al. [73] | Noise Removal Line Segmentation | Structural Features | Rule Based Classifier | Kannada English Hindi | Test Line | TOI, Udyavani Navbharat Times | 99.83 |
| E. Hassan et al. [74] | Adaptive Binarization Skew Detection and Correction Noise Removal | Structural Features | SVM and AdaBoost | Hindi Bangla English | Page text word | Supplementary Books for different courses | 98.76 |
| M.C. Padma et al. [75] | Binarization Noise Removal Skew Correction Morpho-logical Operations | Texture based features | K-NN | Roman Malayalam Telugu Kannada Tamil Urdu | Page | Manually Constructed | 99.68 |
| P.B. Pati et al. [76] | Thresholding noise removal line-word seg | Texture based features | LDC, NN | Hindi Oriya English Tamil | Word | Manually Constructed | 99.7 |

**Table 2.3- Comparative analysis of Offline scripts**

| Researcher | Methodology | | | Type of Script | Granularity | Dataset used | Best accuracy/ performance |
|---|---|---|---|---|---|---|---|
| | Pre-processing | Feature | Classifier | | | | |
| Ptuch R. et al. [1] | Sliding window | 2D image | CNN | English | Line | IAM, RIMES, NIST | 2.46 CER |
| Inkeaw P. et al. [2] | Thresholding noise removal line-word seg | Texture based features | SVM | Asian Languages | Word | THI-C68, NECTEC | 98.52 |
| Cilia N. et al. [3] | Thresholding noise removal line-word segmentation | Stroke based features | HMM, NN | Devanagari, English, Bengali, Tamil | Line | Comparative analysis of algorithms | |
| R. Sarkar et al. [69] | Erosion and Dilation Morphological operation | Topological features | MLP | Devanagari, English Bangla | Word | Sarkar et al. (2010) | 99.29 |
| Raghupathy K. et al. [7] | Bilinear Interpolation | Texture based features | CNN | Tamil | Word | HP Labs Dataset | 97.7 |
| Puri S. et al. [8] | Thresholding noise removal line-word seg | Stroke based features | SVM | Hindi Sanskrit Marathi | Text line | ISIDCHAR V2DMDCHAR | 99.23 |
| Trivedi A. et al. [9] | Thresholding noise removal line-word seg | Stroke based features | CNN Genetic algorithm | Hindi | Word | Indian Statistical Institute (Kolkata) dataset | 98.6 |

## X. CHALLENGES FACED

### *A. Availability of benchmark dataset*

This is a very common problem in research work, though there are plenty of models the lack of standard datasets and database leads to the limitation of comparison between models. The data available for both online and offline handwritten scripts are limited for Indic script thus making the comparison of performance between models difficult. The availability of standard datasets would increase the progress in OSI for the Indic script. The research work in these fields requires a standard dataset or database which satisfies certain criteria. Major of the proposed work has been tested on artificially created datasets and thus no actual comparison can be done between algorithms. For building a system that could be used in real-time the researchers need datasets and databases consisting of handwritten scripts both structured and unstructured to understand the complexity of the scripts and thus training and testing models in real-time by discovering new features.

### B. *Multi-script OCR*

India being a country with more than one language there should be an attempt to make a model that can classify model for identification of script language before classification. The pre-processing for multiple languages may vary from language to language thus there should be an attempt to group pre-processing for multiple similar languages based upon the characteristics. As far as research goes the only model that supports multiple language identification is proposed by Ghosh R. et al. [4] and Bhunia A.K. et al. [6] which can be used to classify and recognize Devanagari and Bengali scripts.

### C. *User-based script identification*

The feature and fonts of handwritten scripts not only differ for languages but also from person to person. The variation of handwriting and script font is influenced by many factors. Beliefs, locality, religion, knowledge of other scripts, education, etc. are some of the factors which determine the features of the handwritten scripts for an individual. It is not possible to incorporate all the variation in the script for each person thus an attempt should be made to that the recognition on the handwritten script based on individuals' previous handwriting when used at the personal or individual level. Research can be conducted for building a model that could be used to tune the classifier for recognition of scripts at an individual level. These researches can be extended to automated models that can learn the features of handwriting and be able to classify the person based on handwriting. These models would surely be able to increase the total output accuracy of general classifiers.

### D. *Feature Selection*

One of the most important steps of optical character recognition and script recognition is feature selection. The selection of the variables depends upon the complexity of the script and the pre-processing methodologies. The general features selection for the online script is Topological features, Spatial spread, directional feature, stoke based feature, and Structural Features. For online script, major features are a time-based and vector-based feature. Some features that can be generated using basic features are velocity-based features for vertical and horizontal selection. Feature selection is a complex task as the whole structure of model selection and pre-processing methodology depends upon it. It must be done from trial and error basis as it plays a key role in the identification of scripts. It is accountable for incorporating all the differences between characters and scripts. Feature selection must be independent and be the same for all scripts. Having a set of roust features can prove helpful in identifying multiple scripts.

## XI. Future work

### A. *Advance pre-processing techniques for old scripts and real-world documents*

The major steps for the pre-procession of offline scripts are usually grey level normalization, binarization, noise removal, thresholding, morphological operation (Erosion Dilation), slant correction, skew correction, etc. Interpolation Smoothing Resampling Normalization skew correction is general pre-processing techniques for online scripts. The main objective of pre-processing and applying filters is to remove the unnecessary noise present in the image and preserve the intended script. Error and noises are very common when a document is scanned or taken the image of because of hardware and human limitations these errors are unavoidable and better pre-processing stages would increase the overall accuracy of the OCR of OSI. All scripts are different and require different pre-processing stages and techniques so to preserve its actual content. This could be done using an agent or a machine learning model to decide the course of pre-processing. Apart from these pre-processing techniques text needs to be segmented and refined before feeding it to the ML model. All these techniques help in obtaining the relevant features for OSI. The choice of proper pre-treatment technologies creates a huge impact for recognition accuracies.

### B. *Feature Selection*

One of the most important step of optical character recognition and script recognition is feature selection. The selection of the variables depends upon the complexity of the script and the pre-processing methodologies. The general features selection for online script are Topological features, Spatial spread, directional feature, stoke based feature and Structural Features. For online script major features are time-based and vector-based feature. Some features for can be generated using basic features are velocity-based features for vertical and horizontal selection. Feature selection is a complex task as the whole structure of model selection and pre-processing methodology depends upon it. It must be done from trial and error basis as it plays a key role in identification of scripts. It is accountable for incorporating all the differences between character and scripts. Feature selection must be independent and be same for all scripts. Having a set of roust feature can prove helpful in identifying multiple scripts.

### C. *Using statistical and corpus-based approach for better prediction*

Statistical analysis and OCR correction can be added to the post-processing stages of OCR. As in generally people write common words and not random alphabets this phase can prove helpful in identification and correction of predicted outcomes. Morphological operation and POS tagging can be used to yield great results. The use of Bi-gram, Tri-gram and n-gram models can boost the overall prediction accuracy.

Validation of multiple classifiers Statistically

Currently there are many classifiers and pattern recognition research models available for study. It cannot be determined by just the best prediction the accuracy and comparative performance of the classifier. To overcome this problem all models can be run multiple times over every dataset and an average of all the performance should be taken into consideration. This method of comparison would surely produce better comparison results among classifier. So that the performance of analysis of the classifier is unbiased the dataset size and division for testing and training must be uniform. Dataset must be trimmed or augmented to satisfy the requirement of the comparison. This method of comparison would provide an actual method for performance analysis between classifiers and this work can be used by researcher further to select and study the models best suited for their problem.

### *D. Ensemble stacking and classifier combination*

As the study proposes that a single model cannot yield perfect results. Thus, future result should target combining the results of multiple classifier to produce better results. Different classifiers take in different feature and thus a variation of feature pool can result to better overall accuracy as it would incorporate different feature that would represent different character. There should be an ensemble voting methodology in which probability of multiple classifiers can accounted. The final prediction of a classifier should be a vector multiplication of all the results vector of each classifier. The multiplication should be biased upon the overall performance of each classifier. There would trade of between accuracy and performance and thus the best suited combination should be tested and applied.

### *E. Visual appearance-based research*

It is evident form the survey that much research has been focused upon textural and stroked based identification of the script. Few researches have been conducted on visual based feature extraction of the script. There has been research on these topics for non – Indic scripts and can be extended to Indic scripts. Classification of Texture and visual appearance is fundamental for many applications in the field of biomedical image processing and recognition, automated visual inspection, remote sensing and content-based image retrieval. These researches can be used for the segregation and the division of characters based on their texture property.

### *F. Future scope of application*

OSI can be combined with many other supervised and on unsupervised models and used in many applications. One the work can be used to aid the un-abled people such as text can be recognized and converted to voice thus adding the deaf. It could be used by people to learn and write different languages. It can have application such as converting documents into searchable text. Researcher can make model that could recognize, and decrypt forgotten old text.

## XII. Conclusion

A comprehensive survey on Indic script recognition and script identification has been done in this review. Script identification and OCR for multilingual scripts is one of the main problems discussed in this survey. Research have been conducted for script identification based on their visual characteristics and structural attributes. According to the proposed work now script identification can be classified into two major categories online and offline script identification. It can also be sub-divided into text, line, word and character-based recognition. This survey has discussed in great details about the pre-processing methodology of different type of script and granularity of classification. Script identification plays a key role in many applications such as reading multiple scripts in which there are use of more than one language, video indexing, sorting of documents and indexing digital library. Identification of the language is initial and the most important step of document processing automation. For script recognition the basic requirement is script recognition for which data about the script is required.

There are many related works in the field of OCR and OSI but still too less work is done on multi-script recognition. Researcher prefer to work on the local problem which belongs to the scope of their region. In many countries the use of a single script is prominent thus resulting in less research on multi-script recognition. India being a country with multiple languages, multi-scripts are used locally. This makes multi-script recognition important in country. Due the variation in languages multi-script identification and recognition using a single module is a difficult task.

The increase in both business transaction and globalization there has been an increase in multi-script identification and classification, due to the increased awareness of the OCR community. Thus, we can see that major work has been done only in the past two decades. Although there has been development in the OCR community major work in Indic script are done on artificially created datasets and the lack of standard datasets result in the unavailability of benchmarking method. Indic script being of very complex in nature needs advance methods and algorithms to account for all their features. Thus, there is a requirement of standardized databases for benchmarking for multiple scripts which provide researcher the resource which is required to develop and test their algorithms

## XIII. References

[1] Ptucha, R., Such, F. P., Pillai, S., Brockler, F., Singh, V., & Hutkowski, P. (2019). Intelligent character recognition using fully convolutional neural networks. Pattern Recognition, 88, 604-613B

[2] Inkeaw, P., Bootkrajang, J., Marukatat, S., Gonçalves, T., & Chaijaruwanich, J. (2019). Recognition of similar characters using gradient features of discriminative regions. Expert Systems with Applications.

[3] Cilia, N. D., De Stefano, C., Fontanella, F., & di Freca, A. S. (2019). A ranking-based feature selection approach for handwritten character recognition. Pattern Recognition Letters, 121, 77-86.

[4] Ghosh, R., Vamshi, C., & Kumar, P. (2019). RNN based online handwritten word recognition in Devanagari and Bengali scripts using horizontal zoning. Pattern Recognition, 92, 203-218.

[5] Choudhury, H., & Prasanna, S. M. (2019). Representation of online handwriting using multi-component sinusoidal model. Pattern Recognition, 91, 200-215.

[6] Bhunia, A. K., Mukherjee, S., Sain, A., Bhunia, A. K., Roy, P. P., & Pal, U. (2020). Indic handwritten script identification using offline-online multi-modal deep network. Information Fusion, 57, 1-14.

[7] Raghupathy, K. B., & Chandrasekaran, S. (2019). Benchmarking on offline Handwritten Tamil Character Recognition using convolutional neural networks. Journal of King Saud University-Computer and Information SciencesI

[8] Puri, S., & Singh, S. P. (2019). An efficient Devanagari character classification in printed and handwritten documents using SVM. Procedia Computer Science, 152, 111-121.

[9] Trivedi, A., Srivastava, S., Mishra, A., Shukla, A., & Tiwari, R. (2018). Hybrid evolutionary approach for Devanagari handwritten numeral recognition using Convolutional Neural Network. Procedia Computer Science, 125, 525-532..

[10] Singh, P. K., Sarkar, R., & Nasipuri, M. (2015). Offline script identification from multilingual Indic-script documents: a state-of-the-art. Computer Science Review, 15, 1-28.

[11] Zamora-Martinez, F., Frinken, V., España-Boquera, S., Castro-Bleda, M. J., Fischer, A., & Bunke, H. (2014). Neural network language models for off-line handwriting recognition. Pattern Recognition, 47(4), 1642-1652.

[12] Poznanski, A., & Wolf, L. (2016). Cnn-n-gram for handwriting word recognition. In Proceedings of the IEEE conference on computer vision and pattern recognition (pp. 2305-2314).

[13] Wigington, C. M. (2018). End-to-end Full-page Handwriting Recognition.

[14] Doetsch, P., Kozielski, M., & Ney, H. (2014, September). Fast and robust training of recurrent neural networks for offline handwriting recognition. In 2014 14th International Conference on Frontiers in Handwriting Recognition (pp. 279-284). IEEE

[15] Almazán, J., Gordo, A., Fornés, A., & Valveny, E. (2014). Word spotting and recognition with embedded attributes. IEEE transactions on pattern analysis and machine intelligence, 36(12), 2552-2566

[16] Barua, S., Malakar, S., Bhowmik, S., Sarkar, R., & Nasipuri, M. (2017). Bangla handwritten city name recognition using gradient-based feature. In Proceedings of the 5th International Conference on Frontiers in Intelligent Computing: Theory and Applications (pp. 343-352). Springer, Singapore

[17] Boughorbel, S., Tarel, J. P., & Boujemaa, N. (2005, September). Generalized histogram intersection kernel for image recognition. In IEEE International Conference on Image Processing 2005 (Vol. 3, pp. III-161). IEEE.

[18] Chaudhary, M., Shikkenawis, G., Mitra, S. K., & Goswami, M. (2012, December). Similar looking Gujarati printed character recognition using Locality Preserving Projection and artificial neural networks. In 2012 Third International Conference on Emerging Applications of Information Technology (pp. 153-156). IEEE.

[19] Cilia, N. D., De Stefano, C., Fontanella, F., & di Freca, A. S. (2019). A ranking-based feature selection approach for handwritten character recognition. Pattern Recognition Letters, 121, 77-86

[20] Gao, T. F., & Liu, C. L. (2008). High accuracy handwritten Chinese character recognition using LDA-based compound distances. Pattern Recognition, 41(11), 3442-3451.

[21] Guyon, I., & Elisseeff, A. (2003). An introduction to variable and feature selection. Journal of machine learning research, 3(Mar), 1157-1182

[22] Xue, B., Zhang, M., Browne, W. N., & Yao, X. (2015). A survey on evolutionary computation approaches to feature selection. IEEE Transactions on Evolutionary Computation, 20(4), 606-626.

[23] Goel, K., Vohra, R., & Bakshi, A. (2014, September). A novel feature selection and extraction technique for classification. In 2014 14th International Conference on Frontiers in Handwriting Recognition (pp. 104-109). IEEE..

[24] Goswami, S., Das, A. K., Chakrabarti, A., & Chakraborty, B. (2017). A feature cluster taxonomy based feature selection technique. Expert Systems with Applications, 79, 76-89.

[25] Guyon, I., & Elisseeff, A. (2003). An introduction to variable and feature selection. Journal of machine learning research, 3(Mar), 1157-1182.

[26] Jaeger, S., Manke, S., Reichert, J., & Waibel, A. (2001). Online handwriting recognition: the NPen++ recognizer. International Journal on Document Analysis and Recognition, 3(3), 169-180.

[27] Graves, A., Liwicki, M., Fernández, S., Bertolami, R., Bunke, H., & Schmidhuber, J. (2008). A novel connectionist system for unconstrained handwriting recognition. IEEE transactions on pattern analysis and machine intelligence, 31(5), 855-868.

[28] Samanta, O., Bhattacharya, U., & Parui, S. K. (2014). Smoothing of HMM parameters for efficient recognition of online handwriting. Pattern Recognition, 47(11), 3614-3629.

[29] Fink, G. A., Vajda, S., Bhattacharya, U., Parui, S. K., & Chaudhuri, B. B. (2010, November). Online Bangla word recognition using sub-stroke level features and hidden Markov models. In 2010 12th International Conference on Frontiers in Handwriting Recognition (pp. 393-398). IEEE

[30] Bharath, A., & Madhvanath, S. (2011). HMM-based lexicon-driven and lexicon-free word recognition for online handwritten Indic scripts. IEEE transactions on pattern analysis and machine intelligence, 34(4), 670-682.

[31] Mandal, S., Prasanna, S. M., & Sundaram, S. (2018). GMM posterior features for improving online handwriting recognition. Expert Systems with Applications, 97, 421-433.

[32] Du, J., Hu, J. S., Zhu, B., Wei, S., & Dai, L. R. (2014, August). A study of designing compact classifiers using deep neural networks for online handwritten Chinese character recognition. In 2014 22nd International Conference on Pattern Recognition (pp. 2950-2955). IEEE.

[33] Kherallah, M., Haddad, L., Alimi, A. M., & Mitiche, A. (2008). On-line handwritten digit recognition based on trajectory and velocity modeling. Pattern Recognition Letters, 29(5), 580-594

[34] Dhieb, T., Ouarda, W., Boubaker, H., Halima, M. B., & Alimi, A. M. (2015, December). Online Arabic writer identification based on beta-elliptic model. In 2015 15th International Conference on Intelligent Systems Design and Applications (ISDA) (pp. 74-79). IEEE.

[35] O'Reilly, C., & Plamondon, R. (2009). Development of a Sigma–Lognormal representation for on-line signatures. Pattern Recognition, 42(12), 3324-3337.

[36] Plamondon, R., O'reilly, C., Galbally, J., Almaksour, A., & Anquetil, É. (2014). Recent developments in the study of rapid human movements with the kinematic theory: Applications to handwriting and signature synthesis. Pattern Recognition Letters, 35, 225-235.

[37] Fischer, A., & Plamondon, R. (2016). Signature verification based on the kinematic theory of rapid human movements. IEEE Transactions on Human-Machine Systems, 47(2), 169-180.

[38] Ferrer, M. A., Diaz, M., Carmona, C. A., & Plamondon, R. (2018). iDeLog: Iterative dual spatial and kinematic extraction of sigma-lognormal parameters. IEEE transactions on pattern analysis and machine intelligence.

[39] Hiremath, P. S., Pujari, J. D., Shivashankar, S., & Mouneswara, V. (2010, February). Script identification in a handwritten document image using texture features. In 2010 IEEE 2nd International Advance Computing Conference (IACC) (pp. 110-114). IEEE.

[40] Chanda, S., Pal, U., & Terrades, O. R. (2009). Word-wise Thai and Roman script identification. ACM Transactions on Asian Language Information Processing (TALIP), 8(3), 11.

[41] Pal, U., Sharma, N., Wakabayashi, T., & Kimura, F. (2007, September). Handwritten numeral recognition of six popular Indian scripts. In Ninth International Conference on Document Analysis and Recognition (ICDAR 2007) (Vol. 2, pp. 749-753). IEEE.

[42] Kumar, B., Bera, A., & Patnaik, T. (2012). Line Based Robust Script Identification for Indian Languages. International Journal of Information and Electronics Engineering, 2(2), 189.

[43] Obaidullah, S. M., Das, S. K., & Roy, K. (2013). A system for handwritten script identification from Indian document. Journal of Pattern Recognition Research, 8(1), 1-12.

[44] Pal, U., & Chaudhuri, B. B. (2004). Indian script character recognition: a survey. pattern Recognition, 37(9), 1887-1899.

[45] Acharya, D. U., Gopakumar, R., & Aithal, P. K. (2010). Multi-Script Line Identification System for Indian Languages. Journal of Computing, 2(11), 107-111.

[46] Handwriting Recognition – "Offline" Approach P. Shankar Rao, J. Aditya {Dept of CSE, Andhra University}

[47] Patel, K., & Gandhi, M. Offline Handwritten Character Recognition: A.

[48] Kannan, R. J., & Prabhakar, R. (2008). An improved handwritten Tamil character recognition system using octal graph.

[49] Razak, Z., Zulkiflee, K., Idris, M. Y. I., Tamil, E. M., Noor, M. N. M., Salleh, R., ... & Yaacob, M. (2008). Off-line handwriting text line segmentation: A review. *International journal of computer science and network security*, *8*(7), 12-20.

[50] Maitra, D. S., Bhattacharya, U., & Parui, S. K. (2015, August). CNN based common approach to handwritten character recognition of

multiple scripts. In *2015 13th International Conference on Document Analysis and Recognition (ICDAR)* (pp. 1021-1025). IEEE.

[51] Shanthi, N., & Duraiswamy, K. (2010). A novel SVM-based handwritten Tamil character recognition system. Pattern Analysis and Applications, 13(2), 173-180.

[52] Jose, T. M., & Wahi, A. (2013). Recognition of Tamil Handwritten Characters using Daubechies Wavelet Transforms and Feed-Forward Backpropagation Network. International Journal of Computer Applications, 64(8).

[53] Sureshkumar, C., & Ravichandran, T. (2010). Handwritten Tamil character recognition and conversion using neural network. International Journal on Computer Science and Engineering, 2(07), 2261-2267.

[54] Bhattacharya, U., Ghosh, S. K., & Parui, S. (2007, September). A two stage recognition scheme for handwritten Tamil characters. In Ninth International Conference on Document Analysis and Recognition (ICDAR 2007) (Vol. 1, pp. 511-515). IEEE.

[55] Boufenar, C., Kerboua, A., & Batouche, M. (2018). Investigation on deep learning for off-line handwritten Arabic character recognition. Cognitive Systems Research, 50, 180-195.

[56] Gaur, A., and Yadav, S. (2015) "Handwritten Hindi character recognition using K-means clustering and SVM", in Fourth International Symposium on Emerging Trends and Technologies in Libraries and Information Services, IEEE Press, pp. 65–70.

[57] Garg, N. K., Kaur, L., and Jndal, M. (2015) "Recognition of offline handwritten Hindi text using middle zone of the words", in Fourteenth International Conference on Computer and Information Science, IEEE Press, pp. 325–328.

[58] Chaudhuri, A., Mandaviya, K., Badelia, P., and Ghosh, S. K. (2017) "Optical character recognition systems for Hindi language", in Optical Character Recognition Systems for Different Languages with Soft Computing: Studies in Fuzziness and Soft Computing, vol. 352, Springer, Cham, pp. 193–216.

[59] Kamble, P. M., and Hegadi, R. S. (2016) "Comparative study of Handwritten Marathi characters recognition based on KNN and SVM classifier", in Santosh K., Hangarge M., Bevilacqua V., and Negi A. (eds) International Conference on Recent Trends in Image Processing and Pattern Recognition: Communications in Computer and Information Science, vol. 709, Springer, Singapore, pp. 93–101

[60] Chaudhuri, A., Mandaviya, K., Badelia, P., and Ghosh, S. K. (2017) "Optical character recognition systems for Hindi language", in Optical Character Recognition Systems for Different Languages with Soft Computing: Studies in Fuzziness and Soft Computing, vol. 352, Springer, Cham, pp. 193–216.

[61] Cortes, C.; Vapnik, V. (1995). "Support-vector networks." Machine Learning. 20 (3): 273-297. doi:10.1007/BF00994018.

[62] Ho, Tin Kam (1995). Random Decision Forests (PDF). Proceedings of the 3rd International Conference on Document Analysis and Recognition, Montreal, QC, 14-16 August 1995. pp. 278-282.

[63] Altman, N. S. (1992). "An introduction to kernel and nearest-neighbor nonparametric regression." The American Statistician. 46 (3): 175 185. doi:10.1080/00031305.1992.10475879.

[64] A. Graves, M. Liwicki, S. Fernandez, R. Bertolami, H. Bunke, J. Schmidhuber. A Novel Connectionist System for Improved Unconstrained andwriting Recognition. IEEE Transactions on Pattern Analysis and Machine Intelligence, vol. 31, no. 5, 2009.

[65] G S Lehal, Nivedan Bhatt, "A Recognition System for Devnagri and English Handwritten Numerals," Proc. Of ICMI, 2000.

[66] Moalla, I., Alimi, A. M., & Benhamadou, A. (2004, September). Extraction of arabic words from multilingual documents. In *Proc. Of Artificial Intelligence and Soft Computing Conference (ASC2004)*.

[67] Ferrer, Miguel A., et al. "Multiple training-one test methodology for handwritten word-script identification." *2014 14th International Conference on Frontiers in Handwriting Recognition*. IEEE, 2014.

[68] Kumar, A., Patnaik, T., & Verma, V. K. (2012). Discrimination of English to other Indian languages (Kannada and Hindi) for OCR system. arXiv preprint arXiv:1205.2164.Sid

[69] Sarkar, R., Das, N., Basu, S., Kundu, M., Nasipuri, M., & Basu, D. K. (2010). Word level script identification from Bangla and Devanagri handwritten texts mixed with Roman script. arXiv preprint arXiv:1002.4007.

[70] Yeotikar, P. P., & Deshmukh, P. R. (2013). Script identification of text words from multilingual Indian document. Int J Comput Appl, 1, 22-29.

[71] Dhanya, D., & Ramakrishnan, A. G. (2002, August). Script identification in printed bilingual documents. In International Workshop on Document Analysis Systems (pp. 13-24). Springer, Berlin, Heidelberg.

[72] Joshi, G. D., Garg, S., & Sivaswamy, J. (2006, February). Script identification from Indian documents. In International Workshop on Document Analysis Systems (pp. 255-267). Springer, Berlin, Heidelberg

[73] Aithal, P. K., Rajesh, G., Acharya, D. U., & Subbareddy, N. K. M. (2010, July). Text line script identification for a tri-lingual document. In 2010 Second International conference on Computing, Communication and Networking Technologies (pp. 1-3). IEEE.

[74] Hassan, E., Garg, R., Chaudhury, S., & Gopal, M. (2011, September). Script based text identification: a multi-level architecture. In Proceedings of the 2011 Joint Workshop on Multilingual OCR and Analytics for Noisy Unstructured Text Data (pp. 1-8).

[75] Padma, M. C., & Vijaya, P. A. (2010). Global approach for script identification using wavelet packet based features. International Journal of Signal Processing, Image Processing and Pattern Recognition, 3(3), 29-40.

[76] Pati, P. B., & Ramakrishnan, A. G. (2008). Word level multi-script identification. Pattern Recognition Letters, 29(9), 1218-1229.

[77] Bhattacharya, U., Plamondon, R., Chowdhury, S. D., Goyal, P., & Parui, S. K. (2017). A sigma-lognormal model-based approach to generating large synthetic online handwriting sample databases. International Journal on Document Analysis and Recognition (IJDAR), 20(3), 155-171.

[78] Liu, C. L., Yin, F., Wang, D. H., & Wang, Q. F. (2011, September). CASIA online and offline Chinese handwriting databases. In 2011 International Conference on Document Analysis and Recognition (pp. 37-41). IEEE

[79] Marti, U. V., & Bunke, H. (2002). The IAM-database: an English sentence database for offline handwriting recognition. International Journal on Document Analysis and Recognition, 5(1), 39-46.

[80] Augustin, E., Carré, M., Grosicki, E., Brodin, J. M., Geoffrois, E., & Prêteux, F. (2006, October). RIMES evaluation campaign for handwritten mail processing. In International Workshop on Frontiers in Handwriting Recognition (IWFHR'06), (pp. 231-235).

[81] Sae-Tang, Sutat, and L. Methasate. "Thai handwritten character corpus." IEEE International Symposium on Communications and Information Technology, 2004. ISCIT 2004.. Vol. 1. IEEE, 2004

[82] Chueaphun, C., Klomsae, A., Marukatat, S., & Chaijaruwanich, J. (2012, December). Lanna Dharma Printed Character Recognition using k-Nearest Neighbor and Conditional Random Fields. In KDIR (pp. 169-174

[83] Surinta, O., Karaaba, M. F., Schomaker, L. R., & Wiering, M. A. (2015). Recognition of handwritten characters using local gradient feature descriptors. Engineering Applications of Artificial Intelligence, 45, 405-414

[84] Viard-Gaudin, C., Lallican, P. M., Knerr, S., & Binter, P. (1999, September). The ireste on/off (ironoff) dual handwriting database. In Proceedings of the Fifth International Conference on Document Analysis and Recognition. ICDAR'99 (Cat. No. PR00318) (pp. 455-458). IEEE.

[85] .Obaidullah, S. M., Halder, C., Santosh, K. C., Das, N., & Roy, K. (2018). PHDIndic_11: page-level handwritten document image dataset of 11 official Indic scripts for script identification. Multimedia Tools and Applications, 77(2), 1643-1678.

[86] http://lipitk.sourceforge.net/datasets/tamilchardata.htm

[87] D'source. History of Devanagari Letterforms. Retrieved on September 10, 2018 from http://www.dsource.in/resource/history-devanagariletterforms/characteristics-script

[88] Bhattacharya, U., & Chaudhuri, B. B. (2005, August). Databases for research on recognition of handwritten characters of Indian scripts. In Eighth International Conference on Document Analysis and Recognition (ICDAR'05) (pp. 789-793). IEEE.

[89] Pati, P. B., & Ramakrishnan, A. G. (2005). Indian script word image dataset.